\documentclass{article}

\usepackage[preprint]{corl_2026} %

\usepackage[T1]{fontenc}
\usepackage{amsmath, amssymb}
\usepackage{graphicx}
\usepackage{booktabs}
\usepackage{multirow}
\usepackage{colortbl}
\usepackage{makecell} 
\usepackage{enumitem}
\usepackage{wrapfig}
\usepackage[linesnumbered,ruled,vlined]{algorithm2e}

\title{DragMesh-2: Physically Plausible Dexterous Hand-Object Interaction with Articulated Objects}

\author{
Tianshan Zhang$^{*}$\quad
Yijia Duan$^{*}$\quad
Yanjun Li$^{*}$\quad
Zeyu Zhang$^{*\dag}$\quad
Hao Tang$^{\ddag}$
\\
[0.3em]
School of Computer Science, Peking University\\
[0.1em]
\footnotesize $^*$Equal contribution.
$^\dag$Project lead.
$^\ddag$Corresponding author: bjdxtanghao@gmail.com.
}

\begin{document}
\maketitle

\begin{abstract}
Dexterous interaction with articulated objects is important for household, assistive, and humanoid manipulation, where multi-finger hands can provide compliant contact patterns beyond parallel-jaw grasping. However, articulated-object manipulation differs from static-object manipulation: the target part cannot be directly actuated, and its motion must emerge through sustained physical hand--handle contact. This makes the transition from object-centric articulated generation to hand-driven dexterous hand--object interaction non-trivial, since geometric trajectory replay or open-loop execution does not model the contact dynamics required to move the articulated part. Moreover, policies trained only for task completion under fixed dynamics can overfit nominal contact loads, especially without tactile or force feedback, and may degrade when the contact load changes. To address these challenges, we present DragMesh-2, a contact-driven framework for dexterous interaction with articulated objects that extends articulated interaction from object-centric generation to hand-driven dexterous hand--object interaction, where articulated motion must arise through physical contact. We further propose PICA, a physically informed contact-aware training mechanism that injects physical signals into policy learning without tactile or force feedback, improving robustness and task success under changing contact loads. Finally, we conduct systematic evaluation across multiple damping conditions and articulated-object categories to study robustness under contact-load variation, and provide a pure-geometry dexterous interaction resource to support future loco-manipulation and humanoid hand--object interaction research. Across seven GAPartNet objects, DragMesh-2 achieves stronger robustness under contact-load variation than the compared methods while maintaining high task success across damping conditions.
Code: \url{https://github.com/AIGeeksGroup/DragMesh-2}.
Website: \url{https://aigeeksgroup.github.io/DragMesh-2}. 
\end{abstract}


\section{Introduction}

Dexterous hand interaction with articulated objects~\cite{okami2024,bao2023dexart,jing2025humanoidgen,lu2025h,lee2025stageact,wang2026paws} is a central problem in robot manipulation and is important in household robotics, assistive systems, and humanoid manipulation settings. Compared with parallel-jaw grippers, dexterous hands provide more compliant multi-finger contact patterns~\cite{liu2023dexrepnet,xu2026contact,dexvla2025}. In recent years, understanding and interacting with articulated objects~\cite{gao2025partrm,liu2025building,le2025articulate,wu2026dipo,wang2025adamanip,wu2022vatmart,Wang2025articubot,zhao2025tac} has become an important research topic in robotics and 3D intelligence. Existing work has mainly focused on articulated structure modeling, motion constraint reasoning, and articulated motion generation. Our previous work, DragMesh~1~\cite{zhang2025dragmesh}, showed that explicit articulation priors can convert user interaction into articulated motion that follows kinematic constraints, enabling object-centric articulated interaction.

However, unlike static objects, articulated objects cannot be directly controlled. Their motion must emerge through sustained hand--object contact, making the transition from object-centric generation to realistic hand--object interaction (HOI) non-trivial. Naive strategies based on geometric trajectory replay, open-loop execution, or direct state control often fail to capture contact dynamics. More importantly, existing reinforcement learning methods are typically trained under fixed dynamics and optimize task completion as the sole objective. Without tactile or force feedback, policies tend to overfit nominal dynamics and rely on dynamics shortcuts rather than learning stable contact behaviors. As a result, a policy that succeeds under nominal damping often degrades rapidly when contact loads change, such as under increased damping. In other words, success under nominal dynamics does not necessarily imply stable contact behavior.

To address these challenges, we propose DragMesh-2, a contact-driven framework for dexterous interaction with articulated objects. DragMesh-2 formulates articulated-object manipulation as a problem that must be completed through real hand--handle interaction: articulated parts cannot be directly controlled, and their motion can only arise through physical interaction between the dexterous hand and the articulated structure. Building on this framework, we further introduce PICA (Physically Informed Contact-Aware) training to improve robustness under changing contact loads without requiring additional force sensing. PICA explicitly injects physically informed signals into policy learning through contact-aware constraints and dynamics randomization, mitigating action saturation, contact detachment, and overfitting to a single dynamics condition. We further combine PICA with temporal contact-response modeling to improve policy representations of changing interaction states. In summary, the main contributions of this work include:

\begin{itemize}[leftmargin=1.2em]

\item We introduce DragMesh-2, a contact-driven framework that extends DragMesh-style articulated interaction from object-centric motion generation to dexterous hand--object interaction. In DragMesh-2, the policy controls only the hand, the target joint has no action channel, and articulated motion must emerge through sustained physical hand--handle contact.

\item  We propose PICA, a physically informed contact-aware training mechanism that injects observable physical signals into policy learning, including contact maintenance, detachment risk, action-boundary regularization, damping variation, and temporal contact response. PICA shifts learning from task-progress-only optimization toward contact-conditioned interaction, improving robustness under contact-load changes without tactile or force feedback.

\item We construct a systematic evaluation protocol across multiple damping conditions and articulated-object categories, together with a pure-geometry dexterous interaction dataset. The protocol evaluates not only task success but also contact-aware diagnostics such as action saturation and detachment, while the generated trajectories provide geometry-guided grasp initialization, task-scale normalization, and tracking references for reproducible contact-driven interaction.
    
\end{itemize}

\section{Related Work}

\textbf{Articulated Object Understanding and Manipulation.}
Articulated objects are common in human environments, but their constrained part motions make perception and manipulation more challenging than rigid-object interaction. Existing studies address articulated object understanding through part-level perception, pose estimation, and shape representation~\cite{geng2023gapartnet,li2020category,mu2021sdf,zhang2026dicart}, joint parameter prediction~\cite{wang2019shape2motion,jain2021screwnet,jiang2022ditto}, and interactive simulation platforms~\cite{makoviychuk2021isaac,xiang2020sapien}. For manipulation, prior methods either infer articulation models for downstream planning~\cite{buchanan2026online} or learn actionable representations and policies from observations~\cite{mo2021where2act,xu2022umpnet,eisner2022flowbot3d}. These efforts, however, mainly target object- or scene-level manipulation with mobile manipulators, grippers, or simplified end-effectors, rather than multi-finger contact-rich interaction with articulated parts. Dexterous interaction with articulated objects remains less explored, as it requires physically compatible coordination between multi-finger contacts and moving object parts.

\textbf{Dexterous Hand-Object Manipulation.}
Classical dexterous manipulation methods modeled multi-finger coordination through contact mechanics, grasp stability, and force closure~\cite{bicchi1995closure,okamura2000overview}. Although theoretically grounded, they require accurate geometry and contact models, which restricts their applicability to diverse objects and uncertain dynamics. Reinforcement learning reduces reliance on explicit modeling by learning high-DoF control policies directly from interaction, and has achieved strong performance on rigid-object in-hand manipulation~\cite{andrychowicz2020learning,rajeswaran2018learning,chen2021system}. Since such policies often require costly exploration, imitation learning and teleoperation-based methods introduce demonstrations and hand-object datasets as behavioral priors~\cite{qin2022dexmv,yang2022oakink,qin2023anyteleop,liu2022hoi4d}. More recent work extends these efforts from rigid-object settings to articulated-object settings, introducing dexterous manipulation benchmarks and hand-object interaction datasets~\cite{bao2023dexart,fan2023arctic}. However, many of these learning pipelines are still primarily evaluated by task progress or success, whereas physically compatible interaction also requires stable contacts, limited interpenetration, and coordinated motion between fingers and articulated parts.

\textbf{Physics-Grounded Manipulation Learning.}
Vision-based teleoperation has enabled dexterous manipulation from visual observations~\cite{handa2020dexpilot,qin2022onehand}, while domain-randomized policy learning has demonstrated autonomous dexterous control without tactile sensing~\cite{andrychowicz2020learning}. In these approaches, contact and dynamics variations are handled through data diversity or domain randomization, without explicitly representing physical variation as a deploy-time adaptation variable. Sim-to-real adaptation methods instead condition policies on dynamics parameters or latent environment factors inferred from recent interaction history during deployment~\cite{yu2017preparing,kumar2021rma}. Such factors are typically low-dimensional and global, making them less suited for capturing the local, state-dependent responses that vary with handle geometry, finger configuration, contact state, and part motion. This motivates PICA to use short-horizon interaction history as a task-level signal for contact-rich articulated manipulation.

Reliable execution of such policies further demands regularized action generation, since saturated joint commands can break contact and destabilize articulated motion. Reward penalties couple action regularization with the task reward and require manual weight tuning. Constrained reinforcement learning~\cite{achiam2017constrained,tessler2019reward} formalizes the separation between task objectives and control constraints through Constrained MDPs and Lagrangian optimization. Motivated by this formulation, PICA introduces separate action-bound and contact-preserving regularization terms alongside the task reward.

\section{Method}
\label{sec:method_main}

\begin{figure*}[t]
\centering
\includegraphics[width=\textwidth]{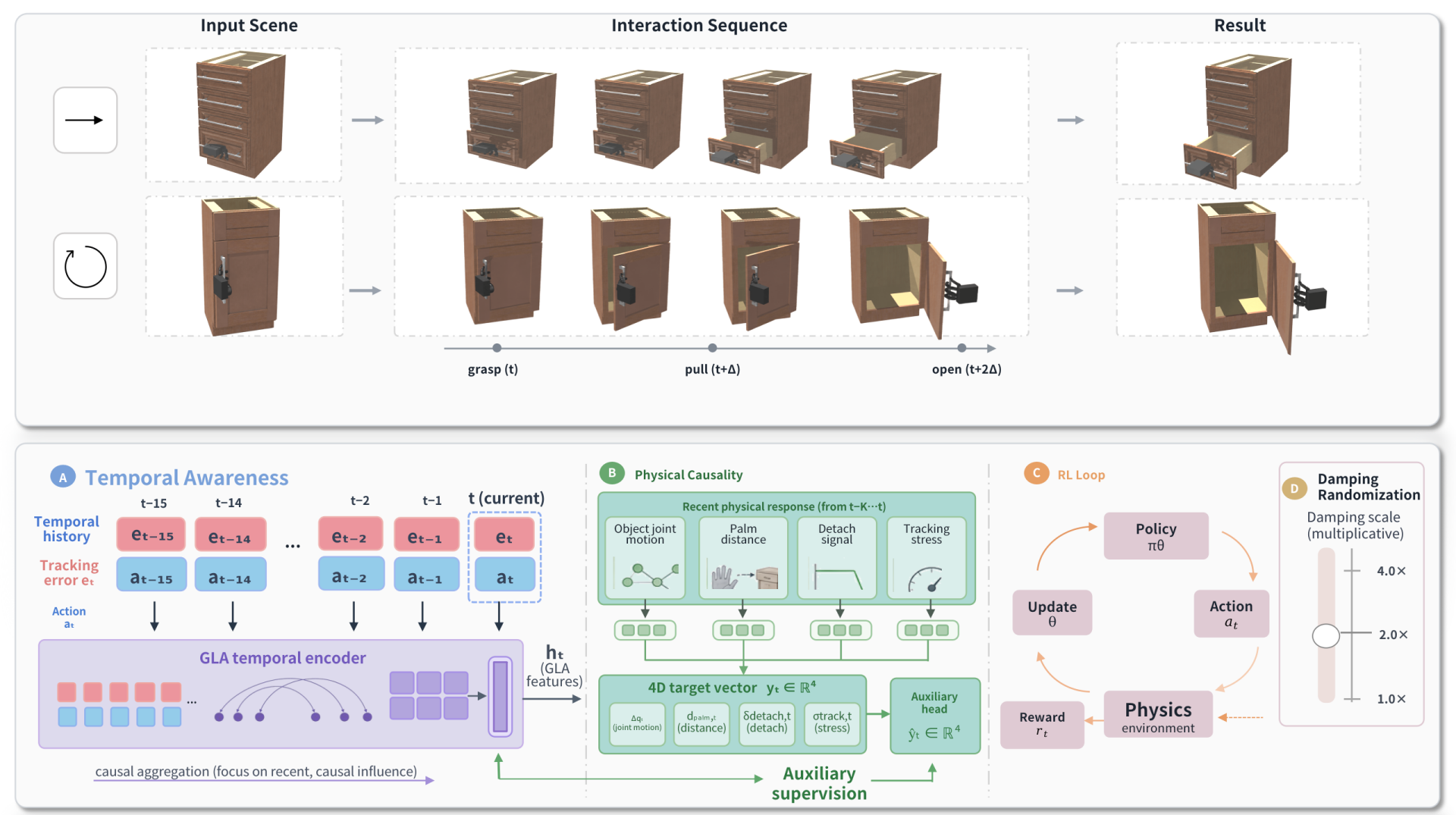}
\caption{Architecture of DragMesh-2.}
\label{fig:framework_overview}
\end{figure*}

\subsection{Contact-Driven Task Formulation}
\label{sec:task_formulation}

DragMesh-2 defines a contact-driven pulling task for a 51-DoF
SMPL-X hand, including 6 virtual wrist DoFs and 45 finger joints.
The policy controls only the hand. The object joint has no action
channel, so the target part can move only through hand--handle
contact. For each GAPartNet object, the target joint is selected as
the object DoF with the largest motion range in a geometry-guided
reference trajectory. The success threshold is set by the relative
motion range of that trajectory,
\begin{equation}
q_{\mathrm{done}} = q_{\min}^{\mathrm{traj}}
+ \rho\,(q_{\max}^{\mathrm{traj}}-q_{\min}^{\mathrm{traj}}),
\label{eq:done_main}
\end{equation}
and task progress is normalized by the same object-specific range,
\begin{equation}
p_t = \max\!\left(0,\,
\frac{q_t^o-q_{\mathrm{start}}}{q_{\mathrm{goal}}-q_{\mathrm{start}}}
\right).
\label{eq:progress_main}
\end{equation}
This definition makes drawers, sliders, and doors comparable without
using a fixed absolute joint displacement.

The observation contains hand joint positions and velocities, handle
pose, relative palm--handle geometry, target-joint state, and
task-scale features derived from progress and remaining distance to
the success threshold. It does not include RGB, depth, point clouds,
force, or tactile signals. The action is a 51-dimensional increment
to the hand PD target, clipped to the joint limits before execution.
The reference trajectory is used only to initialize the expert grasp,
define the target motion scale, and provide a non-learned tracking
baseline; it is not replayed as an object-control command and does
not provide expert action labels.

\subsection{Physically Informed Contact-Aware Learning}
\label{sec:pica}

The reference policy instantiates the PICA signal mechanism with PPO~\cite{schulman2017proximal},
not as a new RL algorithm, but as a controlled policy for the
benchmark. A history token combines the current PD tracking error
and previous action,
\begin{equation}
h_t=[e_t,a_{t-1}],\quad e_t=q_t^{\mathrm{PD}}-q_t^h,
\label{eq:token_main}
\end{equation}
and a GLA encoder\cite{yang2024gated} maps the recent token block to a contact-history
feature. A causal-window auxiliary head predicts observable contact
responses from this feature:
\begin{equation}
y_t = \left[
q_t^o-q_{t-K}^o,\;
\max_{\tau\in[t-K,t]}d_\tau,\;
\mathbb{1}\!\left(\max_{\tau\in[t-K,t]}d_\tau>d_{\mathrm{detach}}\right),\;
\max_{\tau\in[t-K,t]}\lVert e_\tau\rVert_2
\right].
\label{eq:aux_main}
\end{equation}

The four targets describe recent object response, maximum
palm--handle distance, detachment risk, and tracking stress. PICA
incorporates these physical signals in both the environment and policy
levels. At the environment level, the reward augments task progress
with contact maintenance, action regularization, detachment handling,
and successful termination:
\begin{equation}
r_t =
r_{\mathrm{task}}
+ r_{\mathrm{dist}}
+ r_{\mathrm{act}}
+ r_{\mathrm{time}}
+ r_{\mathrm{detach}}
+ r_{\mathrm{success}}
+ r_{\mathrm{bound}}
+ r_{\mathrm{contact}} .
\label{eq:pica_reward_main}
\end{equation}
Here $r_{\mathrm{dist}}$, $r_{\mathrm{bound}}$, and
$r_{\mathrm{contact}}$ encourage contact maintenance and suppress
unsafe separation or saturated control, while the detachment term is
triggered only after the hand has entered and then leaves the effective
contact range.

At the policy level, PICA constrains the temporal representation
through contact-response prediction:
\begin{equation}
\mathcal{L} =
\mathcal{L}_{\mathrm{PPO}}
+ c_v \mathcal{L}_V
+ c_b \mathcal{L}_{\mathrm{bounds}}
+ w_{\mathrm{aux}} \mathcal{L}_{\mathrm{aux}} .
\label{eq:pica_objective_main}
\end{equation}
The auxiliary loss updates the temporal encoder to predict recent
object response, maximum palm--handle distance, detachment risk, and
tracking stress. Thus, PICA augments PPO with observable physical
proxies that bias policy learning away from nominal-success shortcuts
and toward contact-conditioned interaction. Full coefficient values,
network settings, and inference parameters are provided in
Appendix~\ref{sec:framework}.

Evaluation reports task success and progress together with physical
diagnostics. \texttt{clip099} is the fraction of rollout steps whose
maximum action magnitude exceeds $0.99$, and
\texttt{detach\_proxy} is the detachment-failure rate. For damping
set $\mathcal{B}=\{\times1,\times2,\times4\}$ and execution mode
$m\in\{\mathrm{det},\mathrm{stoch}\}$, robustness is summarized as
\begin{equation}
\bar S_m=\frac{1}{|\mathcal{B}|}\sum_{b\in\mathcal{B}}S_{b,m},
\quad
S^{\mathrm{worst}}_m=\min_{b\in\mathcal{B}}S_{b,m}.
\label{eq:robust_main}
\end{equation}
The disaggregated per-damping success values remain important
because averages can hide failure under strong damping.

\subsection{Dataset}
\label{sec:dataset}

The reference contact trajectories are drawn from a dataset that we
generate heuristically, without any learning, directly from GAPartNet
geometry. For each articulated object, a geometry-guided procedure
reads the part, handle, and joint-mobility annotations together with a
SMPL-X hand model, and emits a phased interaction
trajectory---approach, grasp, drag, and release---whose wrist and
finger motion is geometrically consistent with the target joint (full
procedure in Appendix~\ref{sec:trajgen}, Algorithm~\ref{alg:trajectory_generation}).
Each trajectory is stored as a JSON file of per-frame wrist poses and
finger configurations, so the dataset is independent of any policy or
physics backend and can be regenerated for any GAPartNet object that
carries the required annotations.

The dataset contains $277$ trajectories over $7$ GAPartNet categories
(Table~\ref{tab:dataset}). Its distribution follows that of GAPartNet
and is dominated by StorageFurniture, the largest articulated-object
category. Within DragMesh-2 the dataset plays three roles: it
initializes the expert grasp state and target motion scale for the
contact-driven reinforcement-learning task, it defines the non-learned
trajectory-tracking baseline, and we release it as a pure-geometry
interaction resource for future contact-rich and whole-body
loco-manipulation research.

\begin{wraptable}[12]{R}{0.38\textwidth}
    \vspace{-1.0em}
    \centering
    \caption{Heuristic trajectory dataset.}
    \label{tab:dataset}
    \small
    \setlength{\tabcolsep}{4pt}
    \begin{tabular}{lc}
        \toprule
        Category & \#\,Traj. \\
        \midrule
        StorageFurniture & 256 \\
        TrashCan         & 7 \\
        Dishwasher       & 5 \\
        Refrigerator     & 4 \\
        Oven             & 3 \\
        Microwave        & 1 \\
        TableObject           & 1 \\
        \midrule
        Total            & 277 \\
        \bottomrule
    \end{tabular}
    \vspace{-1.0em}
\end{wraptable}

\section{Experiments}
\label{sec:result}
\label{sec:exp_main}

We evaluate on a benchmark of 7 GAPartNet objects spanning three categories (Dishwasher, StorageFurniture, Microwave) and two joint types (5 revolute doors and 2 prismatic drawers). All episodes start from the expert grasp state, and the target part can be opened only through hand--handle contact (Figure~\ref{fig:qual}); the reference contact trajectories that provide this initialization come from our heuristically generated dataset (Section~\ref{sec:dataset}, Contribution~3). Each (method, object, damping, mode) cell uses 20 episodes. Deterministic execution uses the Gaussian mean, while stochastic execution samples from the learned policy. The damping multipliers $\times1$, $\times2$, and $\times4$ measure nominal performance, mild contact-load shift, and strong out-of-distribution (OOD) contact-load shift, respectively.

We compare against a trajectory-tracking replay reference, a GT-part-pose parallel-jaw primitive, and four learned baselines: state-only PPO, flat-history PPO, GRU-PPO, and Transformer-PPO. We further ablate the two physical-structure components of PICA by removing either the physical signals or the GLA temporal encoder. Beyond task success, the contact-aware protocol logs the action-saturation ratio and detachment-failure rate (Appendix~\ref{sec:protocol}); these diagnostics, training hyperparameters, and the full baseline taxonomy are reported in Appendix~\ref{sec:exp}.

\begin{figure*}[t]
\centering
\includegraphics[width=\textwidth]{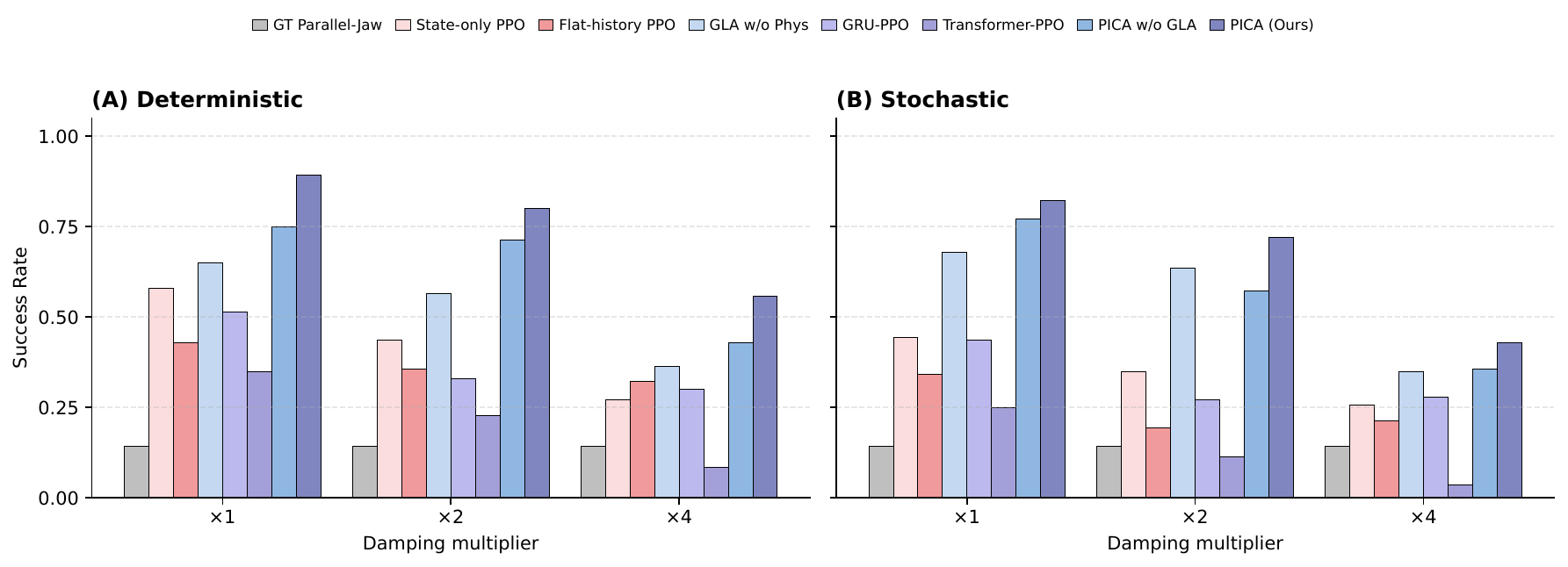}
\caption{Success rate across damping multipliers ($\times 1$, $\times 2$, $\times 4$) under deterministic (A) and stochastic (B) execution, averaged over 7 GAPartNet objects (20 episodes each). We compare a non-learned GT-part-pose parallel-jaw primitive with seven learned methods, including recurrent (GRU-PPO) and Transformer baselines; the primitive is deterministic, so its value is identical across execution modes. Adding physical signals and the GLA encoder yields consistent gains, and PICA attains the highest mean success in all six mode$\times$damping settings and retains the highest absolute success under strong damping ($0.56$ deterministic at $\times4$, versus $0.27$ for state-only PPO and $0.09$ for Transformer-PPO).}
\label{fig:main_grouped}
\end{figure*}

\begin{table*}[t]
\centering
\caption{Per-object success on 7 GAPartNet objects across damping multipliers ($\times1$ nominal, $\times2$ mild, $\times4$ OOD). Each cell is deterministic\,/\,stochastic success (mean over 20 episodes). Trajectory tracking is an open-loop replay reference (deterministic) and the parallel-jaw primitive is deterministic, so both have no stochastic value. The best learned-policy deterministic Avg per damping is in bold.}
\label{tab:main_comparison}
\setlength{\tabcolsep}{4pt}\footnotesize
\resizebox{\textwidth}{!}{%
\begin{tabular}{llcccccccc}
\toprule
\multicolumn{2}{l}{} & \multicolumn{8}{c}{Per-object success (deterministic\,/\,stochastic)} \\
\cmidrule(lr){3-10}
Method & Damp & \shortstack{12583\\{\scriptsize Dishwasher}\\{\scriptsize door}} & \shortstack{45261\\{\scriptsize StorageFurn.}\\{\scriptsize drawer}} & \shortstack{45661\\{\scriptsize StorageFurn.}\\{\scriptsize door}} & \shortstack{45936\\{\scriptsize StorageFurn.}\\{\scriptsize door}} & \shortstack{46440\\{\scriptsize StorageFurn.}\\{\scriptsize drawer}} & \shortstack{48513\\{\scriptsize StorageFurn.}\\{\scriptsize door}} & \shortstack{7310\\{\scriptsize Microwave}\\{\scriptsize door}} & Avg \\
\midrule
\multirow{3}{*}{Traj.\ tracking} & $\times1$ & 1.00 & 1.00 & 1.00 & 1.00 & 1.00 & 1.00 & 1.00 & 1.00 \\
 & $\times2$ & 1.00 & 1.00 & 1.00 & 0.00 & 1.00 & 1.00 & 0.00 & 0.71 \\
 & $\times4$ & 1.00 & 1.00 & 1.00 & 0.00 & 1.00 & 1.00 & 0.00 & 0.71 \\
\midrule
\multirow{3}{*}{Parallel-Jaw} & $\times1$ & 1.00 & 0.00 & 0.00 & 0.00 & 0.00 & 0.00 & 0.00 & 0.14 \\
 & $\times2$ & 1.00 & 0.00 & 0.00 & 0.00 & 0.00 & 0.00 & 0.00 & 0.14 \\
 & $\times4$ & 1.00 & 0.00 & 0.00 & 0.00 & 0.00 & 0.00 & 0.00 & 0.14 \\
\midrule
\multirow{3}{*}{State-only PPO} & $\times1$ & 1.00/0.95 & 0.10/0.20 & 0.00/0.00 & 1.00/0.20 & 0.95/0.90 & 1.00/0.85 & 0.00/0.00 & 0.58/0.44 \\
 & $\times2$ & 0.95/0.50 & 0.10/0.10 & 0.00/0.00 & 0.05/0.10 & 1.00/0.95 & 0.95/0.80 & 0.00/0.00 & 0.44/0.35 \\
 & $\times4$ & 0.00/0.05 & 0.10/0.10 & 0.00/0.00 & 0.00/0.00 & 0.85/0.85 & 0.95/0.80 & 0.00/0.00 & 0.27/0.26 \\
\midrule
\multirow{3}{*}{Flat-history PPO} & $\times1$ & 0.10/0.10 & 1.00/0.90 & 0.05/0.45 & 0.25/0.25 & 0.65/0.20 & 0.00/0.05 & 0.95/0.45 & 0.43/0.34 \\
 & $\times2$ & 0.00/0.00 & 1.00/0.70 & 0.40/0.35 & 0.00/0.00 & 0.65/0.05 & 0.00/0.20 & 0.45/0.05 & 0.36/0.19 \\
 & $\times4$ & 0.00/0.00 & 1.00/0.80 & 0.50/0.30 & 0.00/0.00 & 0.75/0.20 & 0.00/0.20 & 0.00/0.00 & 0.32/0.21 \\
\midrule
\multirow{3}{*}{GRU-PPO} & $\times1$ & 0.60/0.15 & 0.20/0.10 & 1.00/0.90 & 0.00/0.00 & 1.00/1.00 & 0.50/0.75 & 0.30/0.15 & 0.51/0.44 \\
 & $\times2$ & 0.00/0.00 & 0.00/0.05 & 1.00/0.95 & 0.00/0.00 & 1.00/0.85 & 0.00/0.00 & 0.30/0.05 & 0.33/0.27 \\
 & $\times4$ & 0.00/0.00 & 0.00/0.05 & 1.00/1.00 & 0.00/0.00 & 1.00/0.90 & 0.00/0.00 & 0.10/0.00 & 0.30/0.28 \\
\midrule
\multirow{3}{*}{Transformer-PPO} & $\times1$ & 1.00/0.65 & 0.00/0.00 & 0.00/0.00 & 0.00/0.00 & 0.45/0.20 & 0.85/0.70 & 0.15/0.20 & 0.35/0.25 \\
 & $\times2$ & 0.90/0.25 & 0.00/0.05 & 0.00/0.00 & 0.00/0.00 & 0.40/0.25 & 0.15/0.10 & 0.15/0.15 & 0.23/0.11 \\
 & $\times4$ & 0.00/0.00 & 0.00/0.00 & 0.00/0.00 & 0.00/0.00 & 0.60/0.25 & 0.00/0.00 & 0.00/0.00 & 0.09/0.04 \\
\midrule
\multirow{3}{*}{PICA (Ours)} & $\times1$ & 1.00/1.00 & 1.00/1.00 & 0.85/0.90 & 0.90/0.55 & 1.00/0.75 & 0.95/0.85 & 0.55/0.70 & \textbf{0.89}/0.82 \\
 & $\times2$ & 1.00/0.95 & 0.90/0.90 & 0.80/0.70 & 0.65/0.45 & 1.00/0.70 & 0.70/0.80 & 0.55/0.55 & \textbf{0.80}/0.72 \\
 & $\times4$ & 0.95/0.95 & 1.00/0.95 & 0.60/0.65 & 0.25/0.00 & 0.85/0.15 & 0.10/0.30 & 0.15/0.00 & \textbf{0.56}/0.43 \\
\bottomrule
\end{tabular}}
\end{table*}

\paragraph{Main comparison.}

Figure~\ref{fig:main_grouped} and Table~\ref{tab:main_comparison} report the main comparison, and four findings stand out. First, the trajectory-tracking reference reaches $1.00$ deterministic success at $\times1$ on all seven objects, confirming the reference trajectory genuinely drives the target part through contact rather than replaying object states, but its average drops to $0.71$ at $\times2$ and $\times4$ as two objects (45936 and 7310) lose contact once damping rises---open-loop replay alone is not OOD-robust. Second, the GT-part-pose parallel-jaw primitive succeeds on only one object ($0.14$ mean) and is damping-invariant, showing that a geometric primitive cannot substitute for closed-loop dexterous contact control even when the part pose is known. Third, among learned policies PICA attains the best mean in every damping/mode column: deterministic success goes from $0.89$ at $\times1$ to $0.56$ at $\times4$, versus $0.27$ for state-only PPO, $0.32$ for flat-history PPO, $0.30$ for a GRU policy, and $0.09$ for a Transformer policy over the same observation. Fourth, adding richer temporal encoders alone does not close the gap: GRU and Transformer baselines, like the GLA-only ablation (Table~\ref{tab:ablation}, $0.36$ at $\times4$), all lag PICA by at least $0.13$ at $\times4$; the win is therefore the combination of physical signals with temporal contact-response modeling, not the temporal encoder by itself. The per-object numbers also reveal substantial heterogeneity---no single method dominates every instance---which we revisit as a limitation.

\begin{table*}[t]
\centering
\caption{Per-object ablation of the two physical-structure components across damping multipliers. Each cell is deterministic\,/\,stochastic success. \emph{w/o PICA} keeps the GLA temporal encoder but drops the physical signals; \emph{w/o GLA} keeps the physical signals with a flat-history encoder. The best deterministic Avg per damping is in bold.}
\label{tab:ablation}
\setlength{\tabcolsep}{4pt}\footnotesize
\resizebox{\textwidth}{!}{%
\begin{tabular}{llcccccccc}
\toprule
\multicolumn{2}{l}{} & \multicolumn{8}{c}{Per-object success (deterministic\,/\,stochastic)} \\
\cmidrule(lr){3-10}
Method & Damp & \shortstack{12583\\{\scriptsize Dishwasher}\\{\scriptsize door}} & \shortstack{45261\\{\scriptsize StorageFurn.}\\{\scriptsize drawer}} & \shortstack{45661\\{\scriptsize StorageFurn.}\\{\scriptsize door}} & \shortstack{45936\\{\scriptsize StorageFurn.}\\{\scriptsize door}} & \shortstack{46440\\{\scriptsize StorageFurn.}\\{\scriptsize drawer}} & \shortstack{48513\\{\scriptsize StorageFurn.}\\{\scriptsize door}} & \shortstack{7310\\{\scriptsize Microwave}\\{\scriptsize door}} & Avg \\
\midrule
\multirow{3}{*}{w/o PICA (GLA only)} & $\times1$ & 1.00/1.00 & 1.00/1.00 & 0.00/0.40 & 0.35/0.50 & 0.85/0.40 & 0.45/0.70 & 0.90/0.75 & 0.65/0.68 \\
 & $\times2$ & 1.00/1.00 & 1.00/1.00 & 0.05/0.35 & 0.25/0.30 & 0.45/0.40 & 0.40/0.70 & 0.80/0.70 & 0.56/0.64 \\
 & $\times4$ & 0.95/0.75 & 1.00/0.90 & 0.00/0.20 & 0.05/0.05 & 0.00/0.05 & 0.10/0.30 & 0.45/0.20 & 0.36/0.35 \\
\midrule
\multirow{3}{*}{w/o GLA (PICA only)} & $\times1$ & 1.00/1.00 & 0.95/0.85 & 0.00/0.00 & 1.00/0.95 & 1.00/1.00 & 0.45/0.65 & 0.85/0.95 & 0.75/0.77 \\
 & $\times2$ & 1.00/0.85 & 0.85/0.85 & 0.00/0.00 & 1.00/0.70 & 1.00/1.00 & 0.50/0.35 & 0.65/0.25 & 0.71/0.57 \\
 & $\times4$ & 0.05/0.00 & 0.90/0.95 & 0.00/0.00 & 0.80/0.40 & 1.00/1.00 & 0.25/0.15 & 0.00/0.00 & 0.43/0.36 \\
\midrule
\multirow{3}{*}{PICA (Ours)} & $\times1$ & 1.00/1.00 & 1.00/1.00 & 0.85/0.90 & 0.90/0.55 & 1.00/0.75 & 0.95/0.85 & 0.55/0.70 & \textbf{0.89}/0.82 \\
 & $\times2$ & 1.00/0.95 & 0.90/0.90 & 0.80/0.70 & 0.65/0.45 & 1.00/0.70 & 0.70/0.80 & 0.55/0.55 & \textbf{0.80}/0.72 \\
 & $\times4$ & 0.95/0.95 & 1.00/0.95 & 0.60/0.65 & 0.25/0.00 & 0.85/0.15 & 0.10/0.30 & 0.15/0.00 & \textbf{0.56}/0.43 \\
\bottomrule
\end{tabular}}
\end{table*}

\paragraph{Ablation.}
Table~\ref{tab:ablation} isolates the two physical-structure
components of PICA. Using only the GLA temporal encoder (without the
physical signals) reaches $0.65$ deterministic success at $\times1$
and $0.36$ at $\times4$; using only the physical signals (with a
flat-history encoder) reaches $0.75$ and $0.43$; combining the two
reaches $0.89$ and $0.56$. The full model exceeds either component by
at least $0.13$ at $\times4$, and the components help along different
axes---the physical signals contribute more under nominal damping
while the temporal encoder helps more under stochastic
mid-damping---so they are complementary rather than redundant.

\paragraph{Nominal success masks saturation collapse.}
To show that nominal success alone is misleading, we vary only the
training length of the base policy on a single object, before any
contact-stabilization fine-tuning (Table~\ref{tab:trainlen}).
Deterministic nominal ($\times1$) success \emph{rises} from $0.90$ to
$1.00$ as training extends from $150$ to $500$ epochs, yet
strong-damping ($\times4$) success \emph{collapses} from $0.55$ to
$0.10$ while the action-saturation proxy \texttt{clip099} climbs toward
$1.0$. Longer training therefore buys nominal success by driving the
policy into a saturated, low-robustness regime---directly motivating a
protocol that reports OOD damping and saturation alongside success, and
checkpoint selection by OOD robustness rather than by training reward.

\begin{wraptable}[9]{R}{0.46\textwidth}
    \vspace{-1.0em}
    \centering
    \caption{Training-length study of the base policy.}
    \label{tab:trainlen}
    \footnotesize
    \setlength{\tabcolsep}{3pt}
    \begin{tabular}{lccc}
        \toprule
        Base & Succ. $\times1$ & Succ. $\times4$ & \texttt{clip099} $\times4$ \\
        \midrule
        150 ep & 0.90 & \textbf{0.55} & 0.90 \\
        200 ep & 0.90 & 0.50 & 0.97 \\
        300 ep & 1.00 & 0.10 & 0.99 \\
        500 ep & 1.00 & 0.10 & 0.99 \\
        \bottomrule
    \end{tabular}
    \vspace{-1.0em}
\end{wraptable}

\paragraph{Additional studies.}
Appendix~\ref{sec:exp} reports further studies. Extended fine-tuning and damping-range expansion do not yield stable additional OOD gains, indicating that robustness under strong contact load requires richer contact interfaces rather than longer optimization. We also report rollout-level diagnostics and the contact-aware saturation and detachment metrics that motivate the protocol.

\begin{figure}[t]
\centering
\setlength{\tabcolsep}{1.5pt}
\renewcommand{\arraystretch}{0.4}
\begin{tabular}{@{}c ccc@{}}
 & \footnotesize Approach & \footnotesize Grasp & \footnotesize Open \\
\rotatebox[origin=c]{90}{\footnotesize Hardware} &
\includegraphics[width=0.30\linewidth]{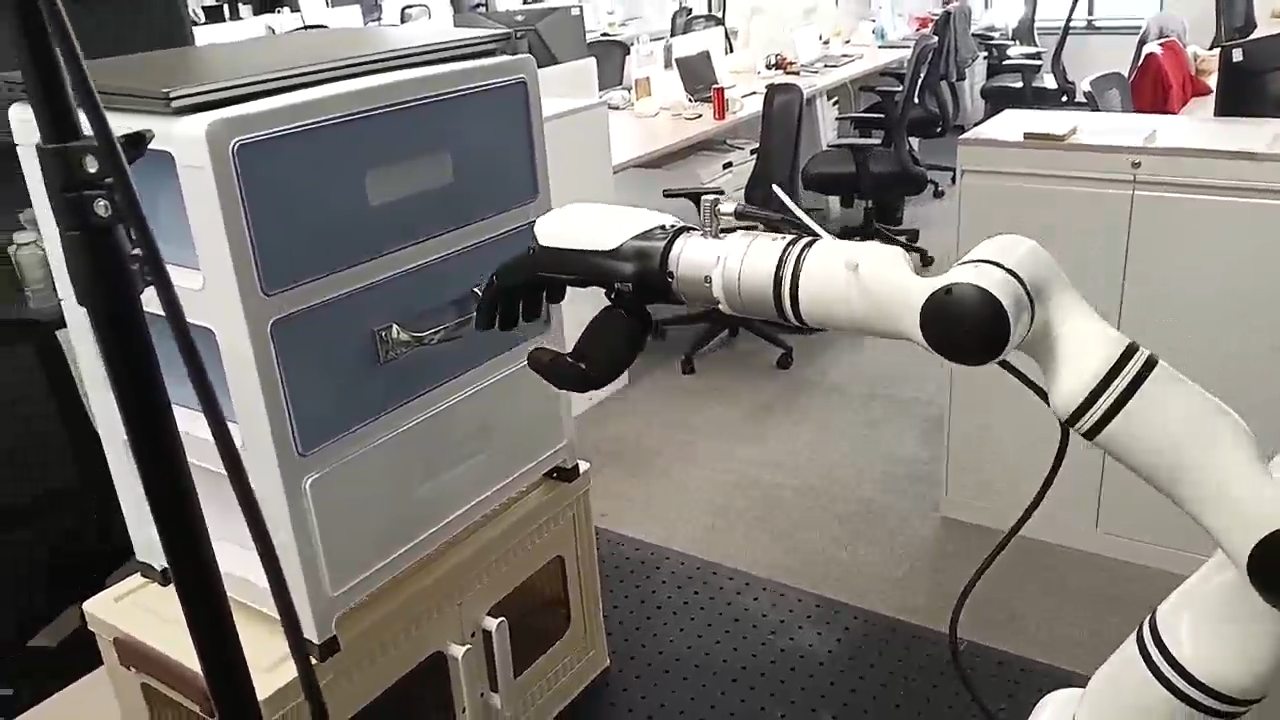} &
\includegraphics[width=0.30\linewidth]{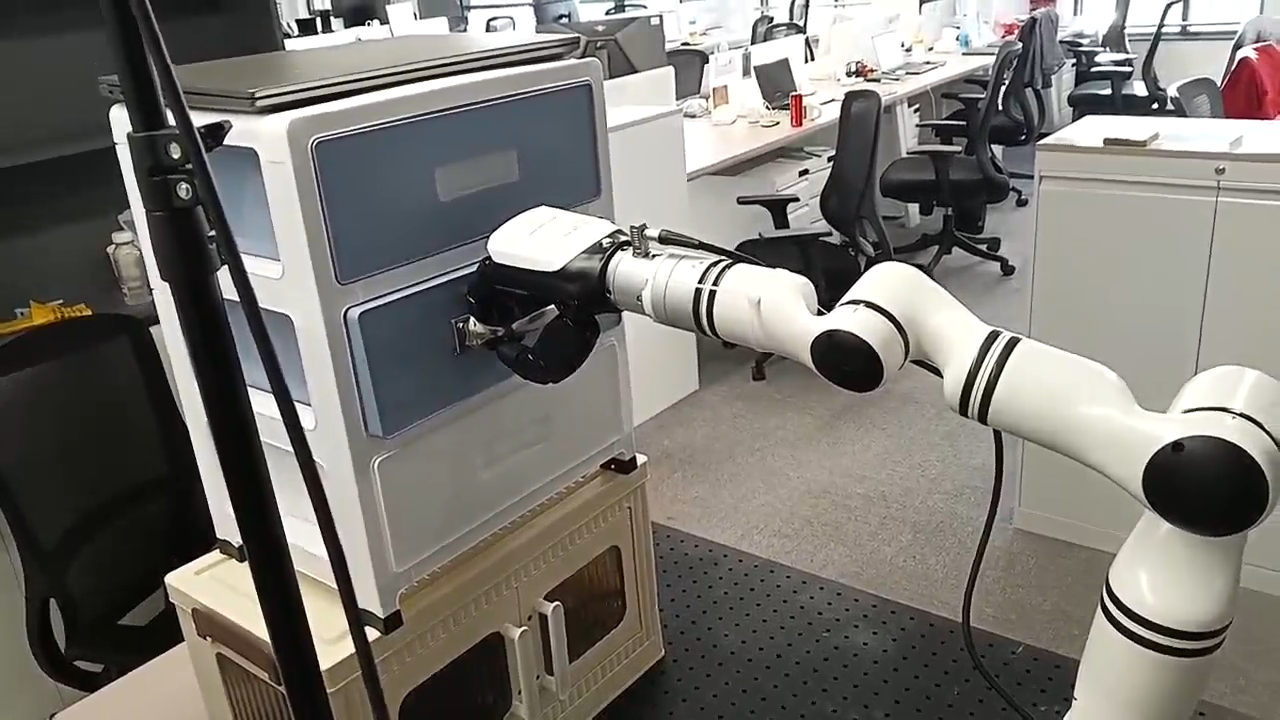} &
\includegraphics[width=0.30\linewidth]{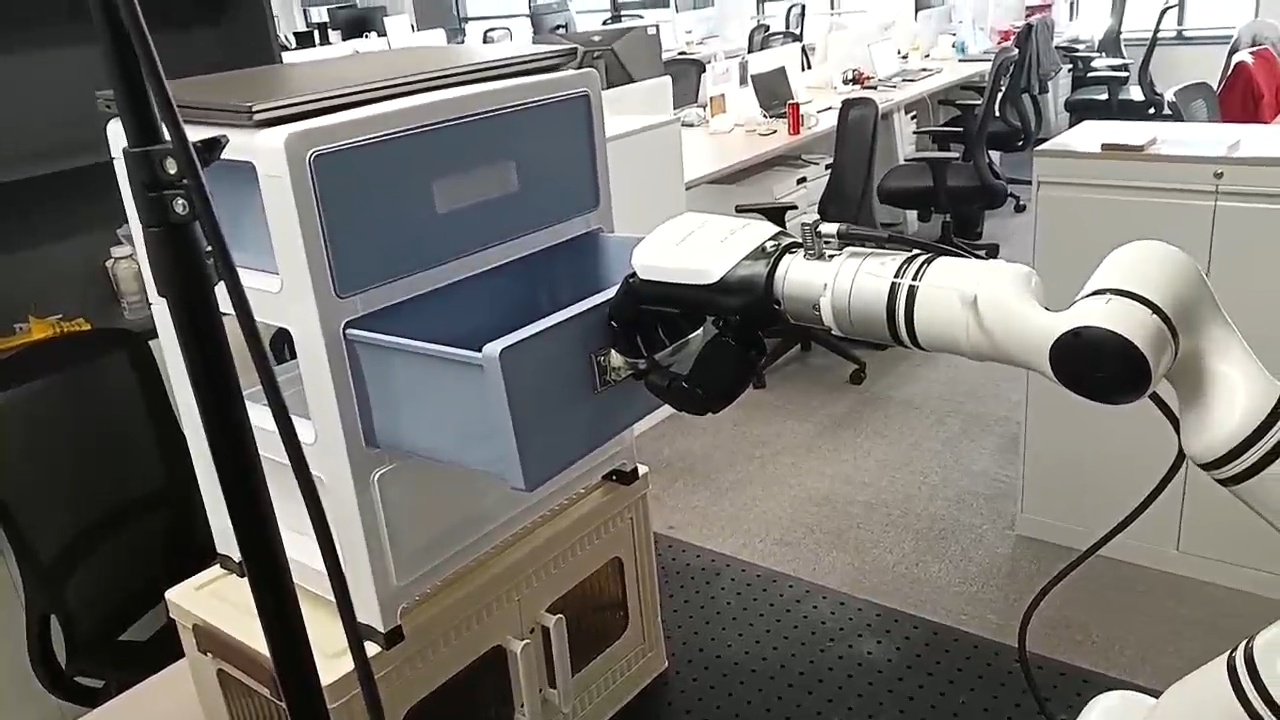} \\
\rotatebox[origin=c]{90}{\footnotesize Simulation} &
\includegraphics[width=0.30\linewidth]{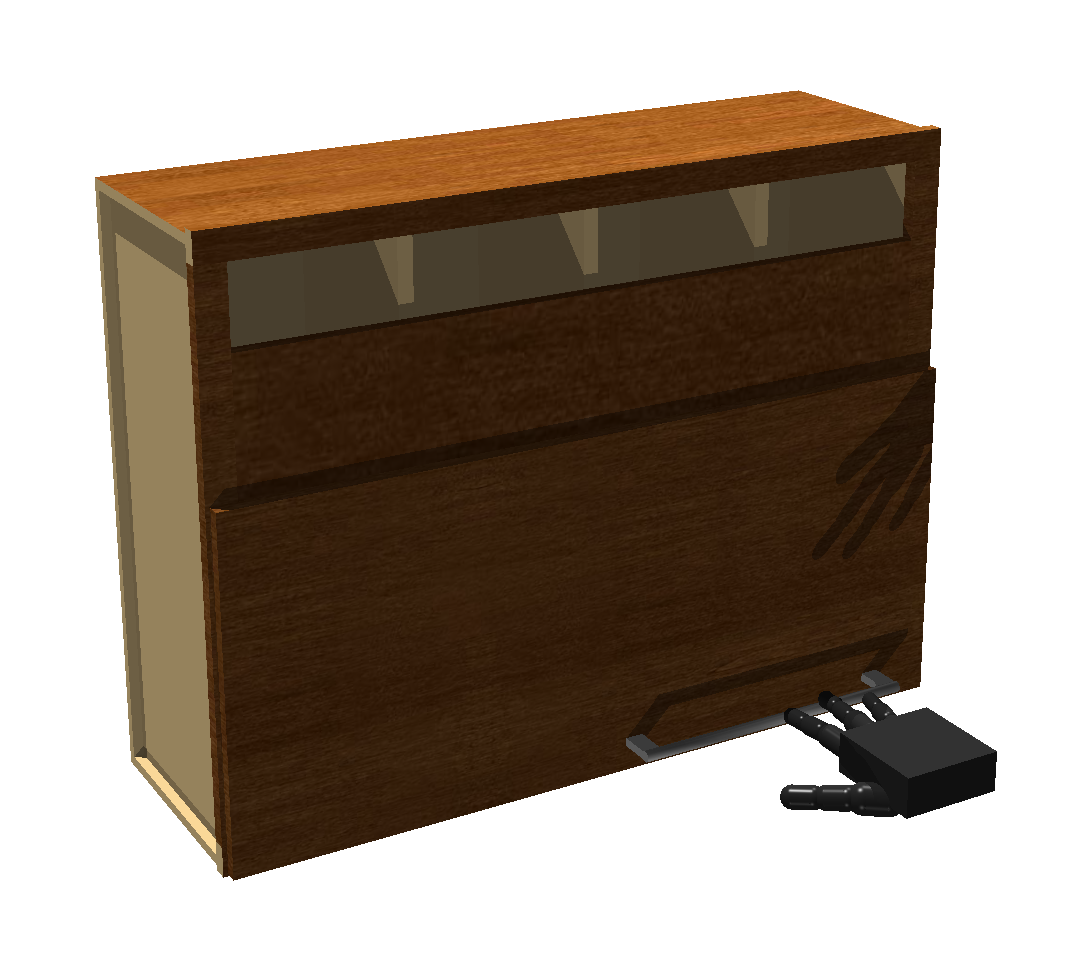} &
\includegraphics[width=0.30\linewidth]{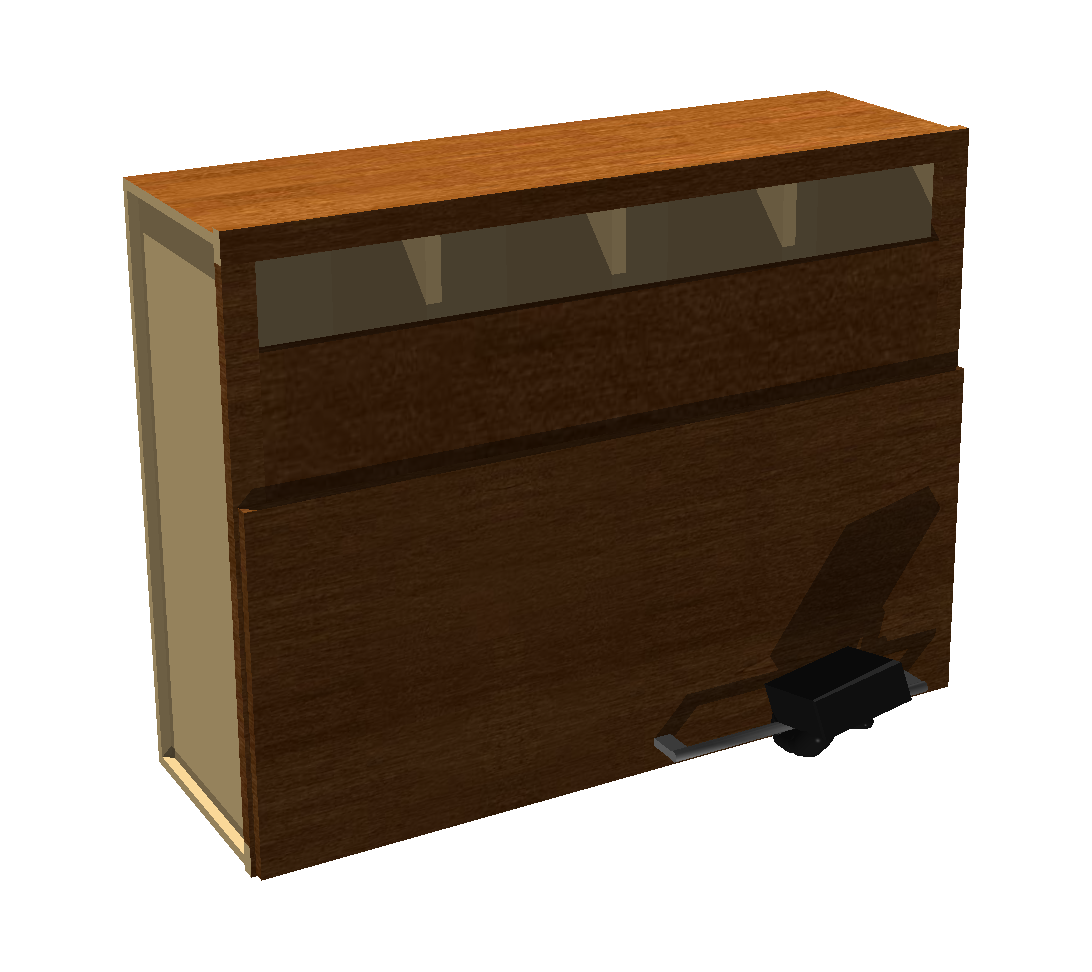} &
\includegraphics[width=0.30\linewidth]{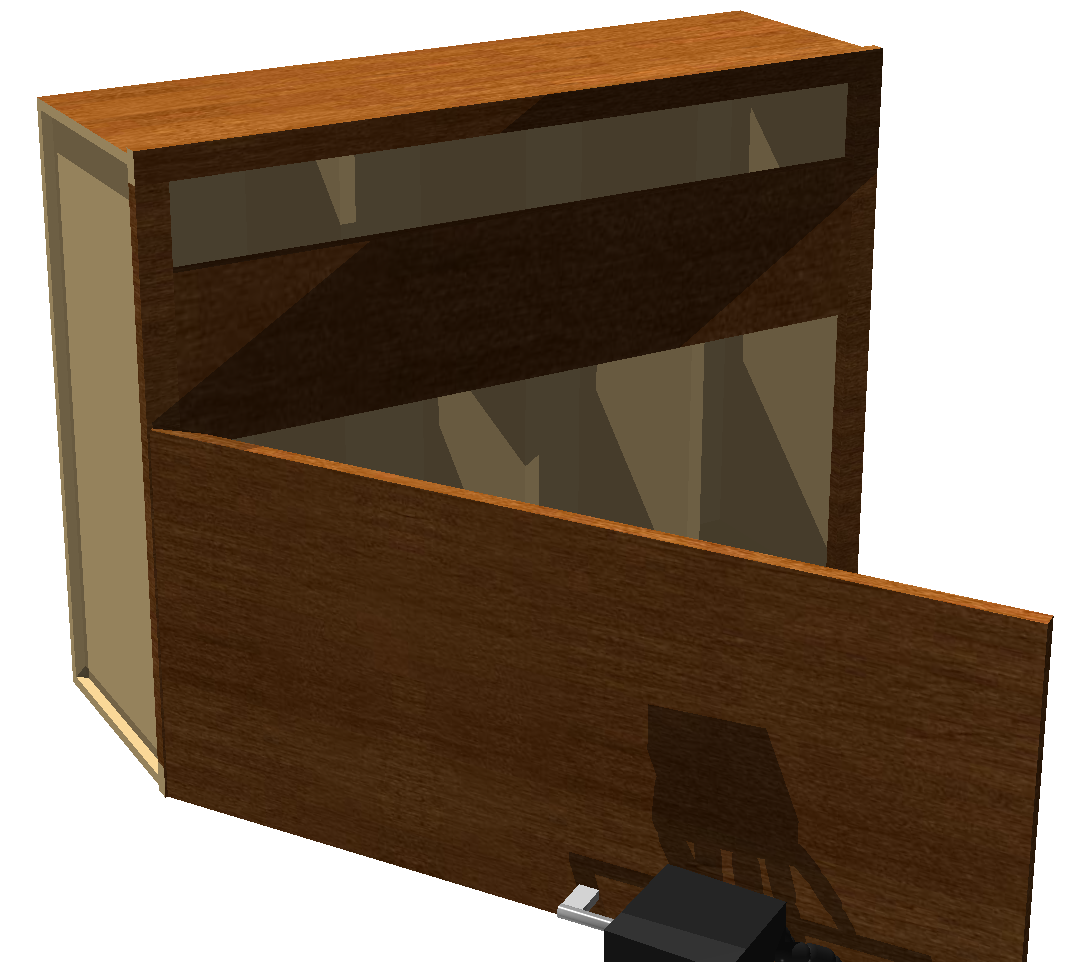} \\
\end{tabular}
\caption{Qualitative illustration of contact-driven articulated interaction.
Top: a hardware feasibility example showing approach, grasp, and opening through hand--handle contact.
Bottom: a simulated rollout of the reference policy on a StorageFurniture instance from the dataset.
The hardware example is included only as a qualitative illustration; all quantitative evaluations are conducted in simulation.}
\label{fig:qual}
\end{figure}

\section{Limitations and Future Work}

DragMesh-2 still has clear limitations, which also point to concrete next steps. First, even with the PICA signals, the learned policy relies on a position-increment action interface and tends toward action saturation under strong contact load: success drops
from $0.89$ at $\times1$ to $0.56$ at $\times4$, and per-object results remain heterogeneous, so no single policy dominates every instance. Because the observation channel provides no force or tactile feedback, contact state can only be inferred indirectly from kinematic error, which appears insufficient for stable light pulling
at high damping. A natural next step is therefore to enrich the contact interface, including wrist force or torque outputs and contact-force or tactile feedback, so that the policy can regulate grip force directly rather than push to the actuator boundary.

Second, our task isolates contact-driven pulling from an expert grasp state and controls a floating dexterous hand. The reference contact
trajectories, however, are full hand--object motion clips that are geometrically consistent with the target joint dynamics, so they extend naturally beyond the hand alone. A promising direction is to couple this upper-body contact interaction with whole-body control, using the dataset as a motion-scale prior for humanoid loco-manipulation, where balance and locomotion must be coordinated with the same physically plausible contact behavior studied here.

\section{Conclusion}
\label{sec:conclusion}

We presented DragMesh-2, a contact-driven framework for dexterous hand--articulated-object interaction, where the target part moves only through physical hand--handle contact. Extending DragMesh~1 from object-centric articulated interaction to hand-driven physical interaction, DragMesh-2 shows that nominal task success does not guarantee stable contact behavior, since policies trained only for task progress can degrade sharply under contact-load shifts. PICA improves robustness by adding physically informed training signals, dynamics randomization, and temporal contact-response modeling without force or tactile feedback. Across seven GAPartNet objects under nominal, moderate, and out-of-distribution damping, DragMesh-2 achieves stronger robustness than competing methods, and we release a pure-geometry interaction dataset for future whole-body loco-manipulation and humanoid HOI.

\clearpage
\acknowledgments{If a paper is accepted, the final camera-ready version will (and probably should) include acknowledgments. All acknowledgments go at the end of the paper, including thanks to reviewers who gave useful comments, to colleagues who contributed to the ideas, and to funding agencies and corporate sponsors that provided financial support.}

\bibliography{example}  %

@String(CVPR    = {IEEE Conf. Comput. Vis. Pattern Recog.})

@String(ICCV    = {Int. Conf. Comput. Vis.})

@String(ECCV    = {Eur. Conf. Comput. Vis.})

@String(NeurIPS = {Adv. Neural Inform. Process. Syst.})

@String(ICML    = {Int. Conf. Mach. Learn.})

@String(ICLR    = {Int. Conf. Learn. Represent.})

@String(CoRL    = {Conf. Robot Learn.})

@String(RSS     = {Robot. Sci. Syst.})

@String(ICRA    = {IEEE Int. Conf. Robot. Autom.})

@String(IROS    = {IEEE/RSJ Int. Conf. Intell. Robots Syst.})

@String(RAL     = {IEEE Robot. Autom. Lett.})

@String(IJRR    = {Int. J. Robot. Res.})

@article{zhang2025dragmesh,
  title={DragMesh: Interactive 3D Generation Made Easy},
  author={Zhang, Tianshan and Zhang, Zeyu and Tang, Hao},
  journal={arXiv preprint arXiv:2512.06424},
  year={2025}
}

@inproceedings{gao2025partrm,
  title={Partrm: Modeling part-level dynamics with large cross-state reconstruction model},
  author={Gao, Mingju and Pan, Yike and Gao, Huan-ang and Zhang, Zongzheng and Li, Wenyi and Dong, Hao and Tang, Hao and Yi, Li and Zhao, Hao},
  booktitle = CVPR,
  year={2025}
}

@article{lu2025h,
  title={{H$^3$DP}: Triply-Hierarchical Diffusion Policy for Visuomotor Learning},
  author={Lu, Yiyang and Tian, Yufeng and Yuan, Zhecheng and Wang, Xianbang and Hua, Pu and Xue, Zhengrong and Xu, Huazhe},
  journal={arXiv preprint arXiv:2505.07819},
  year={2025}
}

@article{jing2025humanoidgen,
        title={HumanoidGen: Data Generation for Bimanual Dexterous Manipulation via LLM Reasoning},
        author={Jing, Zhi and Yang, Siyuan and Ao, Jicong and Xiao, Ting and Jiang, Yugang and Bai, Chenjia},
        journal={arXiv preprint arXiv:2507.00833},
        year={2025}
      }

@inproceedings{okami2024,
    title={OKAMI: Teaching Humanoid Robots Manipulation Skills through Single Video Imitation},
    author={Jinhan Li and Yifeng Zhu and Yuqi Xie and Zhenyu Jiang and Mingyo Seo and Georgios Pavlakos and Yuke Zhu},
    booktitle = CoRL,
    year={2024}
}

@article{lee2025stageact,
  title={StageACT: Stage-Conditioned Imitation for Robust Humanoid Door Opening},
  author={Lee, Moonyoung and Kim, Dong Ki and Bandi, Jai Krishna and Smith, Max and Liao, Aileen and Agha-mohammadi, Ali-akbar and Omidshafiei, Shayegan},
  journal={arXiv preprint arXiv:2509.13200},
  year={2025}
}

@inproceedings{bao2023dexart,
  title={DexArt: Benchmarking Generalizable Dexterous Manipulation with Articulated Objects},
  author={Bao, Chen and Xu, Helin and Qin, Yuzhe and Wang, Xiaolong},
  booktitle=CVPR,
  year={2023}
}

@article{wang2026paws,
  title={PAWS: Perception of Articulation in the Wild at Scale from Egocentric Videos},
  author={Wang, Yihao and Miao, Yang and Zhao, Wenshuai and Yang, Wenyan and Wang, Zihan and Pajarinen, Joni and Van Gool, Luc and Paudel, Danda Pani and Kannala, Juho and Wang, Xi and others},
  journal={arXiv preprint arXiv:2603.25539},
  year={2026}
}

@article{wu2026dipo,
  title={{DIPO}: Dual-State Images Controlled Articulated Object Generation Powered by Diverse Data},
  author={Wu, Ruqi and Wang, Xinjie and Liu, Liu and Guo, Chunle and Qiu, Jiaxiong and Li, Chongyi and Huang, Lichao and Su, Zhizhong and Cheng, Ming-Ming},
  journal=NeurIPS,
  volume={38},
  pages={108665--108689},
  year={2026}
}

@inproceedings{le2025articulate,
  title={{Articulate-Anything}: Automatic Modeling of Articulated Objects via a Vision-Language Foundation Model},
  author={Le, Long and Xie, Jason and Liang, William and Wang, Hung-Ju and Yang, Yue and Ma, Yecheng Jason and Vedder, Kyle and Krishna, Arjun and Jayaraman, Dinesh and Eaton, Eric},
  booktitle=ICLR,
  year={2025}
}

@article{dexvla2025,
  title={End-to-End Dexterous Arm-Hand VLA Policies via Shared Autonomy: VR Teleoperation Augmented by Autonomous Hand VLA Policy for Efficient Data Collection},
  author={Cui, Yu and Zhang, Yujian and Tao, Lina and Li, Yang and Yi, Xinyu and Li, Zhibin},
  journal={arXiv preprint arXiv:2511.00139},
  year={2025}
}

@inproceedings{xu2026contact,
  title     = {Contact-Grounded Policy: Dexterous Visuotactile Policy with Generative Contact Grounding},
  author    = {Xu, Zhengtong and Wang, Yeping and Abbatematteo, Ben and Preechayasomboon, Jom and Chan, Sonny and Colonnese, Nick and Memar, Amirhossein H.},
  booktitle = RSS,
  year      = {2026}
}

@article{schulman2017proximal,
  title={Proximal policy optimization algorithms},
  author={Schulman, John and Wolski, Filip and Dhariwal, Prafulla and Radford, Alec and Klimov, Oleg},
  journal={arXiv preprint arXiv:1707.06347},
  year={2017}
}

@inproceedings{yang2024gated,
  title={Gated linear attention transformers with hardware-efficient training},
  author={Yang, Songlin and Wang, Bailin and Shen, Yikang and Panda, Rameswar and Kim, Yoon},
  booktitle=ICML,
  year={2024}
}

@article{zhao2025tac,
  title={Tac-{M}an: Tactile-Informed Prior-Free Manipulation of Articulated Objects},
  author={Zhao, Zihang and Li, Yuyang and Li, Wanlin and Qi, Zhenghao and Ruan, Lecheng and Zhu, Yixin and Althoefer, Kaspar},
  journal={IEEE Transactions on Robotics (T-RO)},
  year={2025},
}

@inproceedings{Wang2025articubot,
      title={ArticuBot: Learning Universal Articulated Object Manipulation Policy via Large Scale Simulation},
      author={Wang, Yufei and Wang, Ziyu and Nakura, Mino and Bhowal, Pratik and Kuo, Chia-Liang and Chen, Yi-Ting and Erickson, Zackory and Held, David},
      booktitle=RSS,
      year={2025}}

@inproceedings{
wu2022vatmart,
title={{VAT}-Mart: Learning Visual Action Trajectory Proposals for Manipulating 3D {ART}iculated Objects},
author={Ruihai Wu and Yan Zhao and Kaichun Mo and Zizheng Guo and Yian Wang and Tianhao Wu and Qingnan Fan and Xuelin Chen and Leonidas Guibas and Hao Dong},
booktitle=ICLR,
year={2022},
}

@inproceedings{wang2025adamanip,
    title={AdaManip: Adaptive Articulated Object Manipulation Environments and Policy Learning},
    author={Wang, Yuanfei and Zhang, Xiaojie and Wu, Ruihai and Li, Yu and Shen, Yan and Wu, Mingdong and He, Zhaofeng and Wang, Yizhou and Dong, Hao},
    booktitle=  ICLR,
    year={2025},
  }

@inproceedings{liu2023dexrepnet,
  title={Dexrepnet: Learning dexterous robotic grasping network with geometric and spatial hand-object representations},
  author={Liu, Qingtao and Cui, Yu and Ye, Qi and Sun, Zhengnan and Li, Haoming and Li, Gaofeng and Shao, Lin and Chen, Jiming},
  booktitle=IROS,
  year={2023},
}

@inproceedings{liu2025building,
  title={Building Interactable Replicas of Complex Articulated Objects via Gaussian Splatting},
  author={Liu, Yu and Jia, Baoxiong and Lu, Ruijie and Ni, Junfeng and Zhu, Song-Chun and Huang, Siyuan},
  booktitle = ICLR,
  year={2025},
}

@inproceedings{geng2023gapartnet,
  title     = {{GAPartNet}: Cross-Category Domain-Generalizable Object Perception and Manipulation via Generalizable and Actionable Parts},
  author    = {Geng, Haoran and Xu, Helin and Zhao, Chengyang and Xu, Chao and Yi, Li and Huang, Siyuan and Wang, He},
  booktitle = CVPR,
  year      = {2023}
}

@inproceedings{li2020category,
  title     = {Category-Level Articulated Object Pose Estimation},
  author    = {Li, Xiaolong and Wang, He and Yi, Li and Guibas, Leonidas J. and Abbott, A. Lynn and Song, Shuran},
  booktitle = CVPR,
  year      = {2020}
}

@inproceedings{mu2021sdf,
  title     = {{A-SDF}: Learning Disentangled Signed Distance Functions for Articulated Shape Representation},
  author    = {Mu, Jiteng and Qiu, Weichao and Kortylewski, Adam and Yuille, Alan and Vasconcelos, Nuno and Wang, Xiaolong},
  booktitle = ICCV,
  year      = {2021}
}

@article{zhang2026dicart,
  title   = {{DICArt}: Advancing Category-Level Articulated Object Pose Estimation in Discrete State-Spaces},
  author  = {Zhang, Li and Mei, Mingyu and Wang, Ailing and Meng, Xianhui and Zhong, Yan and Song, Xinyuan and Liu, Liu and Wang, Rujing and He, Zaixing and Lu, Cewu},
  journal = {arXiv preprint arXiv:2602.19565},
  year    = {2026}
}

@inproceedings{wang2019shape2motion,
  title     = {{Shape2Motion}: Joint Analysis of Motion Parts and Attributes from {3D} Shapes},
  author    = {Wang, Xiaogang and Zhou, Bin and Shi, Yahao and Chen, Xiaowu and Zhao, Qinping and Xu, Kai},
  booktitle = CVPR,
  year      = {2019}
}

@inproceedings{jain2021screwnet,
  title     = {{ScrewNet}: Category-Independent Articulation Model Estimation from Depth Images Using Screw Theory},
  author    = {Jain, Ajinkya and Lioutikov, Rudolf and Chuck, Caleb and Niekum, Scott},
  booktitle = ICRA,
  year      = {2021}
}

@inproceedings{jiang2022ditto,
  title     = {{Ditto}: Building Digital Twins of Articulated Objects from Interaction},
  author    = {Jiang, Zhenyu and Hsu, Cheng-Chun and Zhu, Yuke},
  booktitle = CVPR,
  year      = {2022}
}

@article{buchanan2026online,
  title={Online Estimation and Manipulation of Articulated Objects},
  author={Buchanan, Russell and R{\"o}fer, Adrian and Moura, Jo{\~a}o and Valada, Abhinav and Vijayakumar, Sethu},
  journal={arXiv preprint arXiv:2601.01438},
  year={2026}
}

@inproceedings{makoviychuk2021isaac,
  title     = {{Isaac Gym}: High Performance {GPU}-Based Physics Simulation for Robot Learning},
  author    = {Makoviychuk, Viktor and Wawrzyniak, Lukasz and Guo, Yunrong and Lu, Michelle and Storey, Kier and Macklin, Miles and Hoeller, David and Rudin, Nikita and Allshire, Arthur and Handa, Ankur and State, Gavriel},
  booktitle = {NeurIPS Datasets and Benchmarks},
  year      = {2021}
}

@inproceedings{xiang2020sapien,
  title     = {{SAPIEN}: A SimulAted Part-Based Interactive ENvironment},
  author    = {Xiang, Fanbo and Qin, Yuzhe and Mo, Kaichun and Xia, Yikuan and Zhu, Hao and Liu, Fangchen and Liu, Minghua and Jiang, Hanxiao and Yuan, Yifu and Wang, He and Yi, Li and Chang, Angel X. and Guibas, Leonidas J. and Su, Hao},
  booktitle = CVPR,
  year      = {2020}
}

@inproceedings{eisner2022flowbot3d,
  title     = {{FlowBot3D}: Learning {3D} Articulation Flow to Manipulate Articulated Objects},
  author    = {Eisner, Ben and Zhang, Harry and Held, David},
  booktitle = RSS,
  year      = {2022}
}

@article{bicchi1995closure,
  title   = {On the Closure Properties of Robotic Grasping},
  author  = {Bicchi, Antonio},
  journal = IJRR,
  volume  = {14},
  number  = {4},
  pages   = {319--334},
  year    = {1995}
}

@inproceedings{okamura2000overview,
  title     = {An Overview of Dexterous Manipulation},
  author    = {Okamura, Allison M. and Smaby, Niels and Cutkosky, Mark R.},
  booktitle = ICRA,
  year      = {2000}
}

@article{andrychowicz2020learning,
  title   = {Learning Dexterous In-Hand Manipulation},
  author  = {Andrychowicz, Marcin and Baker, Bowen and Chociej, Maciek and J{\'o}zefowicz, Rafal and McGrew, Bob and Pachocki, Jakub and Petron, Arthur and Plappert, Matthias and Powell, Glenn and Ray, Alex and Schneider, Jonas and Sidor, Szymon and Tobin, Josh and Welinder, Peter and Weng, Lilian and Zaremba, Wojciech},
  journal = IJRR,
  volume  = {39},
  number  = {1},
  pages   = {3--20},
  year    = {2020}
}

@inproceedings{rajeswaran2018learning,
  title     = {Learning Complex Dexterous Manipulation with Deep Reinforcement Learning and Demonstrations},
  author    = {Rajeswaran, Aravind and Kumar, Vikash and Gupta, Abhishek and Vezzani, Giulia and Schulman, John and Todorov, Emanuel and Levine, Sergey},
  booktitle = RSS,
  year      = {2018}
}

@inproceedings{chen2021system,
  title     = {A System for General In-Hand Object Re-Orientation},
  author    = {Chen, Tao and Xu, Jie and Agrawal, Pulkit},
  booktitle = CoRL,
  year      = {2021}
}

@inproceedings{qin2022dexmv,
  title     = {{DexMV}: Imitation Learning for Dexterous Manipulation from Human Videos},
  author    = {Qin, Yuzhe and Wu, Yueh-Hua and Liu, Shaowei and Jiang, Hanwen and Yang, Ruihan and Fu, Yang and Wang, Xiaolong},
  booktitle = ECCV,
  year      = {2022}
}

@inproceedings{yang2022oakink,
  title     = {{OakInk}: A Large-Scale Knowledge Repository for Understanding Hand-Object Interaction},
  author    = {Yang, Lixin and Li, Kailin and Zhan, Xinyu and Wu, Fei and Xu, Anran and Liu, Liu and Lu, Cewu},
  booktitle = CVPR,
  year      = {2022}
}

@inproceedings{qin2023anyteleop,
  title     = {{AnyTeleop}: A General Vision-Based Dexterous Robot Arm-Hand Teleoperation System},
  author    = {Qin, Yuzhe and Yang, Wei and Huang, Binghao and Van Wyk, Karl and Su, Hao and Wang, Xiaolong and Chao, Yu-Wei and Fox, Dieter},
  booktitle = RSS,
  year      = {2023}
}

@inproceedings{liu2022hoi4d,
  title     = {{HOI4D}: A {4D} Egocentric Dataset for Category-Level Human-Object Interaction},
  author    = {Liu, Yunze and Liu, Yun and Jiang, Che and Lyu, Kangbo and Wan, Weikang and Shen, Hao and Liang, Boqiang and Fu, Zhoujie and Wang, He and Yi, Li},
  booktitle = CVPR,
  year      = {2022}
}

@inproceedings{fan2023arctic,
  title     = {{ARCTIC}: A Dataset for Dexterous Bimanual Hand-Object Manipulation},
  author    = {Fan, Zicong and Taheri, Omid and Tzionas, Dimitrios and Kocabas, Muhammed and Kaufmann, Manuel and Black, Michael J. and Hilliges, Otmar},
  booktitle = CVPR,
  year      = {2023}
}

@inproceedings{handa2020dexpilot,
  title     = {{DexPilot}: Vision-Based Teleoperation of Dexterous Robotic Hand-Arm System},
  author    = {Handa, Ankur and Van Wyk, Karl and Yang, Wei and Liang, Jacky and Chao, Yu-Wei and Wan, Qian and Birchfield, Stan and Ratliff, Nathan D. and Fox, Dieter},
  booktitle = ICRA,
  year      = {2020}
}

@article{qin2022onehand,
  title   = {From One Hand to Multiple Hands: Imitation Learning for Dexterous Manipulation from Single-Camera Teleoperation},
  author  = {Qin, Yuzhe and Su, Hao and Wang, Xiaolong},
  journal = RAL,
  volume  = {7},
  number  = {4},
  pages   = {10873--10881},
  year    = {2022}
}

@inproceedings{kumar2021rma,
  title     = {{RMA}: Rapid Motor Adaptation for Legged Robots},
  author    = {Kumar, Ashish and Fu, Zipeng and Pathak, Deepak and Malik, Jitendra},
  booktitle = RSS,
  year      = {2021}
}

@inproceedings{achiam2017constrained,
  title     = {Constrained Policy Optimization},
  author    = {Achiam, Joshua and Held, David and Tamar, Aviv and Abbeel, Pieter},
  booktitle = ICML,
  year      = {2017}
}

@inproceedings{tessler2019reward,
  title     = {Reward Constrained Policy Optimization},
  author    = {Tessler, Chen and Mankowitz, Daniel J. and Mannor, Shie},
  booktitle = ICLR,
  year      = {2019}
}

@inproceedings{yu2017preparing,
  title     = {Preparing for the Unknown: Learning a Universal Policy with Online System Identification},
  author    = {Yu, Wenhao and Tan, Jie and Liu, C. Karen and Turk, Greg},
  booktitle = RSS,
  year      = {2017}
}

@inproceedings{mo2021where2act,
  title     = {{Where2Act}: From Pixels to Actions for Articulated {3D} Objects},
  author    = {Mo, Kaichun and Guibas, Leonidas J. and Mukadam, Mustafa and Gupta, Abhinav and Tulsiani, Shubham},
  booktitle = ICCV,
  year      = {2021}
}

@article{xu2022umpnet,
  title   = {{UMPNet}: Universal Manipulation Policy Network for Articulated Objects},
  author  = {Xu, Zhenjia and He, Zhanpeng and Song, Shuran},
  journal = RAL,
  volume  = {7},
  number  = {2},
  pages   = {2447--2454},
  year    = {2022}
}

\clearpage
\appendix

\section{Additional Method Details}
\label{sec:framework}

This appendix gives implementation-level details omitted from the
main text. Section~\ref{sec:method_main} contains the task
definition, core PICA signals, and evaluation metrics; the appendix
expands only the observation and control representation, reward
terms, temporal encoder, auxiliary supervision, optimization
settings, additional diagnostics, and implementation parameters.

\subsection{Observation and Control Details}
\label{sec:obs}

Because the environment does not provide explicit force or tactile
sensors, DragMesh~2 is intrinsically a partially observable Markov
decision process. The kinematic state at a single frame is
insufficient to infer the contact impedance of the hand--object
system, load variation, or potential detachment trends. The protocol
therefore uses a state-only observation that describes the visible
geometric and joint state at the current frame, while the policy
approximates the implicit physical state from recent control
history. No RGB, depth, point cloud, or semantic segmentation input
is used.

The state at time $t$ contains hand joint positions and velocities,
the handle pose, the relative hand--handle geometry, the object
joint state, and task-scale features derived from the target joint
position:
\begin{equation}
s_t = \left[
q_t^h,\,\dot q_t^h,\,
x_t^{\mathrm{handle}},\,r_t^{\mathrm{handle}},\,
x_t^{\mathrm{handle}}-x_t^{\mathrm{palm}},\,
d_t,\,
q_t^o,\,\dot q_t^o,\,
\phi(q_t^o)
\right].
\label{eq:state}
\end{equation}
Here $q_t^h,\dot q_t^h$ are 51-dimensional hand joint positions and
velocities, $x_t^{\mathrm{handle}},r_t^{\mathrm{handle}}$ are the
handle position and orientation, $d_t$ is the distance from palm
center to handle center, and $q_t^o,\dot q_t^o$ are the object joint
position and velocity. The task-scale features $\phi(q_t^o)$ contain
relative task progress, the remainder to the success threshold, and
the object-level motion scale $\Delta q =
q_{\mathrm{goal}}-q_{\mathrm{start}}$. These features are
deterministic functions of $q_t^o$ and the task boundary defined by
the trajectory. Keeping them explicit helps learn a value function
shared across doors, drawers, and sliders with different motion
ranges.

The policy outputs a 51-dimensional continuous action $a_t$ for
incremental control of the virtual wrist joint and finger joints.
The action is first clipped to $[-1,1]$ and then scaled by $\alpha$
into a local increment of the hand PD target:
\begin{equation}
\Delta q_t^h = \alpha a_t,\quad
q_{t,\mathrm{target}}^h =
\mathrm{clip}(q_t^h+\Delta q_t^h,
\mathbf{q}_{\min}^h,\mathbf{q}_{\max}^h).
\label{eq:action}
\end{equation}
The target is sent to a position PD controller, while the object
joint is affected only through simulated contact. Since no policy
channel directly drives the object joint, the target part can only be
opened through contact between the hand and the handle. This incremental
action representation reduces the difficulty of high-dimensional
hand control while preserving the role of contact dynamics in
producing object motion. Action scaling, control frequency, and the
inference-time execution mode are listed in the appendix.

\subsection{Evaluation-Protocol Details}
\label{sec:protocol}

The main text defines the contact-aware metrics and robustness
summary. The appendix records the only additional protocol detail:
the non-learned trajectory-tracking baseline feeds each next-frame
hand pose from the reference trajectory as the hand PD target, while
the object joint state is never replayed. The target part is still
driven only through hand--handle contact, so this baseline tests
whether the reference motion can induce physical opening rather than
reproduce stored object states.

\subsection{Reference Policy: Physical Signal Mechanism (PICA)}
\label{sec:pica_app}

\subsubsection{Physical-Plausibility Reward}
\label{sec:reward}

The reference policy treats contact maintenance and action regularity
as differentiable training targets. In addition to the task-progress
reward, the policy is subject to contact maintenance, action
magnitude, and termination constraints. The palm--handle distance
$d_t$ defines a weak contact
maintenance term:
\begin{equation}
r_{\mathrm{dist}} = -w_{\mathrm{dist}}\,d_t
+ w_{\mathrm{near}}\exp(-\kappa d_t).
\end{equation}
The task-progress term is the per-step increment of the target joint:
\begin{equation}
r_{\mathrm{task}} = w_{\mathrm{task}}\,(q_t^o - q_{t-1}^o).
\end{equation}
The action and time costs are
\begin{equation}
r_{\mathrm{act}} = -w_{\mathrm{act}}\,
\mathrm{mean}(a_t^2),\quad
r_{\mathrm{time}} = -w_{\mathrm{time}}.
\end{equation}
If the palm has previously approached the handle and the palm--handle
distance later exceeds $d_{\mathrm{detach}}$ before the task is
done, the episode is declared a detachment failure and incurs a
one-shot penalty; if the target joint reaches the success threshold
a one-shot success bonus is granted. The detachment criterion only
triggers after the policy has been within effective contact range,
which acts as a contact gate. The policy must remain on the contact
manifold and cannot evade later action costs by releasing contact or
moving away from the handle. Success, detachment failure, and
exceeding the maximum episode length each terminate the episode.

To suppress brittle pulling driven by saturated actions, the
reference policy adds action-boundary and contact-distance
regularizers:
\begin{equation}
r_{\mathrm{bound}} = -w_{\mathrm{bound}}\,
\mathrm{mean}\!\left(
\max(|a_t|-a_{\mathrm{sat}},0)^2\right),
\end{equation}
\begin{equation}
r_{\mathrm{contact}} = -w_{\mathrm{contact}}\,
\max(d_t-d_{\mathrm{safe}},0)^2.
\end{equation}
The total reward is
\begin{equation}
r_t = r_{\mathrm{dist}} + r_{\mathrm{task}} + r_{\mathrm{act}}
+ r_{\mathrm{time}} + r_{\mathrm{detach}}
+ r_{\mathrm{success}}
+ r_{\mathrm{bound}} + r_{\mathrm{contact}}.
\label{eq:totalreward}
\end{equation}
The four explicit physical constraints, namely saturation gating
$r_{\mathrm{bound}}$, contact-distance regularization
$r_{\mathrm{contact}}$, detachment gating $r_{\mathrm{detach}}$, and
the damping randomization described next, together constitute the
PICA signal mechanism. The reward signal of the policy is driven not
only by nominal task progress, but also by whether the task is
completed under contact-maintaining and action-regularized
conditions. Coefficient values are listed in the appendix.

For fine-tuning, two optional contact-stabilization rewards serve as
ablation modules. The first, ARAM, is an adaptive version of
$r_{\mathrm{bound}}$; in high-impedance stalled states, it imposes
additional penalties on high-magnitude actions, so the policy cannot
bypass contact resistance through sustained saturated pulling. The
second, Reconfig, encourages small-amplitude hand reconfiguration
when the policy stalls in contact, allowing the policy to
re-establish effective contact rather than persist with a failing
pulling pose. These two modules are used in the diagnostic ablations
as the ARAM, Reconfig, and combined Both fine-tunes; they do not
change the DragMesh~2 task definition, the observation interface, or
the evaluation protocol.

\subsubsection{Damping Randomization}
\label{sec:damprand}

Damping randomization tests whether the policy relies on a single
nominal dynamics setting. At each environment reset, a damping scale
for the target object joint is drawn uniformly from a specified
interval and applied to the nominal damping. This perturbation
exposes the policy during training to pulling responses under
varying resistance, reducing dependence on a single dynamics
setting. The default training interval is $[1.0, 2.0]$; friction
randomization is not used. Evaluation uses higher damping
multipliers, including $\times 4$, to construct OOD dynamics tests.

\subsubsection{Contact-History Temporal Encoder}
\label{sec:gla}

Contact-rich dexterous manipulation is strongly time-dependent. The
hand pose and object joint state at a single frame do not fully
reveal whether contact is stable, whether the hand is sliding off the
handle, or whether recent actions are producing object response. The
reference policy therefore includes recent control history in the
policy state. Each history token consists of the hand PD tracking
error and the previous action:
\begin{equation}
h_t = [e_t,\,a_{t-1}],\quad
e_t = q_t^{\mathrm{PD}} - q_t^h.
\label{eq:token}
\end{equation}
The full history block is
\begin{equation}
H_t = [h_{t-L+1},\,\ldots,\,h_t].
\label{eq:block}
\end{equation}
This sequence reflects observable physical response during contact.
When the hand is impeded by reaction forces from the handle, the
tracking error grows. When strong control produces no target-joint
progress, the action--error pattern in the history exposes high
impedance or invalid contact. When the palm drifts away from the
handle, the change in contact distance combined with the action
response signals impending detachment.

The reference policy uses a two-branch actor--critic architecture.
The state $s_t$ is encoded by an MLP, while the history block $H_t$
is projected by a linear layer into a token representation and fed
to a Gated Linear Attention temporal encoder. Compared with a
history-concatenated MLP or standard self-attention, the gating in
GLA is more sensitive to abrupt phase transitions in contact
dynamics, such as transient impacts or sliding. By default, the
output corresponding to the last history token is used as the
contact-history feature:
\begin{equation}
z_t^{\mathrm{hist}} = \mathrm{GLA}(H_t)_L.
\label{eq:zhist}
\end{equation}
After normalization, this temporal feature is concatenated with the
current state feature to form the fused representation $z_t$. The
actor head outputs the Gaussian policy mean $\mu_t$ from $z_t$, and
the critic head outputs the state value $V(s_t, H_t)$ from the same
fused representation. Network widths, head counts, history length,
and policy standard deviation are listed in the appendix.

\subsubsection{Causal-Window Contact-Response Auxiliary Supervision}
\label{sec:aux}

A temporal encoder does not automatically learn a representation
consistent with contact dynamics: under a training loop driven only
by task reward, it can equally well degrade into a shortcut
representation under the nominal dynamics. The reference policy
therefore applies causal-window physical auxiliary supervision to
$z_t^{\mathrm{hist}}$, requiring this feature to recover observable
contact-response signals from recent history. These signals serve as
an implicit time-domain representation of contact impedance: they do
not measure contact forces directly, but they characterize how the
hand--object system responds to control inputs through object
response, contact distance, and hand tracking error.

The environment maintains a buffer of the most recent $K+1$ physics
steps of palm--handle distance and target object joint position. The
auxiliary head receives only $z_t^{\mathrm{hist}}$ and predicts
\begin{equation}
y_t = \left[
q_t^o-q_{t-K}^o,\;
\max_{\tau\in[t-K,t]}d_\tau,\;
\mathbb{1}\!\left(\max_{\tau\in[t-K,t]}d_\tau>d_{\mathrm{detach}}\right),\;
\max_{\tau\in[t-K,t]}\lVert e_\tau\rVert_2
\right].
\label{eq:aux}
\end{equation}
The four channels denote, respectively, the recent object joint
response, the maximum palm--handle distance in the window, the
detachment-risk proxy, and the maximum tracking stress in the
causal window. Here $e_\tau = q_\tau^{\mathrm{PD}} - q_\tau^h$, and
$\max_{\tau\in[t-K,t]}\lVert e_\tau\rVert_2$ measures the largest
tracking residual between PD target and hand state in the window,
which can be read, in a compliant-control sense, as an observable
proxy for contact load. The auxiliary targets are passed only to the
auxiliary head as training supervision. When the actor, critic, and
GLA backbone process the observation, these target channels are
explicitly removed from the input and never serve as additional
state features for policy decisions.

DragMesh~2 uses causal windows rather than single-step differences
to construct auxiliary targets. Single-step physical quantities in
rigid-body contact are susceptible to high-frequency noise from the
contact solver, and single-step transients cannot represent the slow
degradation of contact quality, such as gradual palm slip before a
failure flag is triggered. Windowed object response, maximum contact
distance, and the detachment-risk proxy provide a smoother and more
stable impedance representation, making the auxiliary supervision a
more reliable constraint for guiding the temporal encoder toward
contact-response representations.

The auxiliary loss uses a channel-weighted mean-squared error:
\begin{equation}
\mathcal{L}_{\mathrm{aux}} =
\omega_q\,\ell(q_{\mathrm{response}})
+ \omega_d\,\ell(d_{\max})
+ \omega_b\,\ell(b_{\mathrm{detach}})
+ \omega_e\,\ell(E_{\mathrm{track}}).
\label{eq:laux}
\end{equation}
This loss updates both the auxiliary head and the GLA temporal
encoder. Its purpose is not to reward any specific action, but to
encourage the temporal encoder to represent whether actions produce
object response, whether contact is maintained, and whether the hand
is under tracking stress. The actor and critic therefore receive a
temporally encoded representation with explicit physical meaning.
The channel weights and the window length are listed in the
appendix.

\subsubsection{Policy Optimization}
\label{sec:opt}

The reference policy is optimized with PPO using GAE for advantage
estimation. PPO serves as the optimization vehicle rather than a
contribution. The training objective combines the clipped policy
loss, the value loss, the actor output-boundary loss, and the
physical auxiliary loss:
\begin{equation}
\mathcal{L} = \mathcal{L}_{\mathrm{PPO}}
+ c_v\,\mathcal{L}_V
+ c_b\,\mathcal{L}_{\mathrm{bounds}}
+ w_{\mathrm{aux}}\,\mathcal{L}_{\mathrm{aux}}.
\label{eq:opt}
\end{equation}
$\mathcal{L}_{\mathrm{bounds}}$ acts directly on the actor output
distribution and discourages the policy mean from approaching the
action boundary, while the environment-side $r_{\mathrm{bound}}$
penalizes saturated actions after execution.
The auxiliary weight $w_{\mathrm{aux}}$ follows a linear warmup, so
the policy first acquires basic control through the task reward and
is then constrained by contact-response prediction. All training
hyperparameters, reward coefficients, action scaling, history
length, and inference modes are listed in the appendix.

The components in the reference-policy layer, namely the
contact-regularized reward, damping randomization, GLA
contact-history encoding, and causal-window auxiliary supervision,
jointly instantiate the PICA signal mechanism. The training
objective explicitly contains observable physical proxies, the
temporal representation is constrained toward implicit states
consistent with contact response, and the policy is trained to
behave stably under varied contact loads. This instantiation is one
possible implementation; the specific network structure, reward
coefficients, and optimization hyperparameters are not the main
contribution of this paper. They serve to test whether the defined
task and protocol can distinguish contact-conditioned behavior from
unstable shortcut behavior.

\section{Reference-Trajectory Generation and Dataset Construction}
\label{sec:trajgen}

This section details the geometry-guided procedure that produces the reference contact trajectories and the released dataset (Section~\ref{sec:dataset}); it uses only GAPartNet geometry and annotations, with no learning.

\paragraph{Target-part selection.}
For a given GAPartNet object, the system selects the target part
according to the semantic annotation: parts whose category contains
\texttt{slider} or \texttt{drawer} are preferred; otherwise
\texttt{door} is chosen; if neither matches, the last annotated part
is used as a fallback. The handle annotation nearest the front face
of the target part is then chosen; its oriented bounding box gives the
center $\mathbf{c}_h$, long axis $\mathbf{l}_h$, outward normal
$\mathbf{n}_h$, short axis $\mathbf{s}_h$, and thickness $d_h$. If
no explicit handle annotation exists, the front face of the target
part is used as a fallback to construct the contact geometry.

\paragraph{Wrist and finger poses.}
The wrist orientation $\mathbf{R}_w$ is constructed from
$\mathbf{l}_h$ and $\mathbf{n}_h$ so that the palm faces the
interaction region and the fingers wrap along the long axis of the
handle. Since control is anchored to the wrist coordinate frame
while contact occurs near the palm, the wrist position compensates
for the palm-center offset $\mathbf{o}_{\mathrm{palm}}$:
\begin{equation}
\mathbf{x}_w = \mathbf{c}_h - \mathbf{R}_w \mathbf{o}_{\mathrm{palm}}.
\end{equation}
The finger configurations include an open pose
$\mathbf{q}_{\mathrm{open}}$, a pre-grasp pose
$\mathbf{q}_{\mathrm{pre}}$, and a force-closure grasp pose
$\mathbf{q}_{\mathrm{grasp}}$, with $\mathbf{q}_{\mathrm{grasp}}$
adjusted to the handle thickness $d_h$.

\paragraph{Approach phase.}
The pre-grasp position is offset along the outward normal:
\begin{equation}
\mathbf{x}_{\mathrm{pre}} = \mathbf{x}_w + \delta \mathbf{n}_h,
\end{equation}
and the wrist is linearly interpolated from $\mathbf{x}_{\mathrm{pre}}$
to $\mathbf{x}_w$ over $T_{\mathrm{approach}}$ steps, while the
fingers move from $\mathbf{q}_{\mathrm{open}}$ to
$\mathbf{q}_{\mathrm{pre}}$:
\begin{equation}
\mathbf{x}_t = (1-\alpha_t)\mathbf{x}_{\mathrm{pre}}
+ \alpha_t \mathbf{x}_w,\quad
\mathbf{q}_t = (1-\alpha_t)\mathbf{q}_{\mathrm{open}}
+ \alpha_t \mathbf{q}_{\mathrm{pre}},
\end{equation}
with $\alpha_t = t/T_{\mathrm{approach}}$.

\paragraph{Grasp phase.}
The wrist pose is held fixed and the fingers close from
$\mathbf{q}_{\mathrm{pre}}$ to $\mathbf{q}_{\mathrm{grasp}}$.

\paragraph{Drag phase.}
For a prismatic joint, the interaction center translates along the
outward normal of the handle:
\begin{equation}
\mathbf{c}_t = \mathbf{c}_h + \alpha_t d \mathbf{n}_h,
\end{equation}
with $d$ a preset drag distance. For a revolute joint, the
interaction center rotates about the joint axis $\mathbf{a}_j$ and
the joint origin $\mathbf{o}_j$:
\begin{equation}
\mathbf{c}_t = \mathbf{o}_j + \mathbf{R}(\alpha_t\theta,
\mathbf{a}_j)(\mathbf{c}_h - \mathbf{o}_j),
\end{equation}
where $\theta$ is a preset rotation angle. The wrist position is
recomputed from the updated interaction center at each step; in the
revolute case, the wrist orientation rotates synchronously so the
palm tracks the handle.

\paragraph{Release phase.}
After the drag completes, the fingers gradually open to
$\mathbf{q}_{\mathrm{open}}$ and the wrist retracts along the final
outward normal.

The full pseudocode is given in
Algorithm~\ref{alg:trajectory_generation}. The generator has no
learned parameters and depends only
on the GAPartNet geometry and mobility annotations, so it runs
directly on any articulated object equipped with both geometry and
mobility annotations. The phase
durations $T_{\mathrm{approach}}$, $T_{\mathrm{grasp}}$,
$T_{\mathrm{drag}}$, $T_{\mathrm{release}}$, the geometric offset
$\delta$, the drag distance $d$, and the rotation angle $\theta$
are set adaptively to the object type.

\begin{algorithm}[t]
\caption{Geometry-Guided Interaction Trajectory Generation}
\label{alg:trajectory_generation}
\KwIn{Articulated object $o$; GAPartNet annotations $\mathcal{A}$;
      mobility annotations $\mathcal{M}$; hand model $\mathcal{H}$;
      motion parameters $\Theta = \{\delta, d, \theta,
      T_{\mathrm{approach}}, T_{\mathrm{grasp}},
      T_{\mathrm{drag}}, T_{\mathrm{release}}\}$.}
\KwOut{Reference contact trajectory
      $\tau = \{(\mathbf{x}_t,\mathbf{R}_t,
      \mathbf{q}_t,\mathbf{s}_t^{\mathrm{obj}})\}_{t=1}^{T}$.}

\BlankLine
\tcc{Scene and annotation initialization.}
Initialize a physics scene with object $o$ and hand model $\mathcal{H}$\;
Load part bounding boxes, categories, link names from $\mathcal{A}$
and joint information from $\mathcal{M}$\;

\BlankLine
\tcc{Target part, handle and mobility selection.}
$p^\star \leftarrow \textsc{SelectTargetPart}(\mathcal{A})$\;
$h^\star \leftarrow \textsc{FindAssociatedHandle}(p^\star,\mathcal{A})$\;
$(\mathbf{c}_h,\mathbf{l}_h,\mathbf{n}_h,\mathbf{s}_h,d_h)
   \leftarrow \textsc{DecomposeHandleBBox}(h^\star)$\;
$m \leftarrow \textsc{QueryMobilityType}(p^\star,\mathcal{M})$\;

\BlankLine
\tcc{Wrist pose and hand shape.}
$\mathbf{R}_w \leftarrow
   \textsc{ComputeWristOrientation}(\mathbf{l}_h,\mathbf{n}_h)$\;
$\mathbf{x}_w \leftarrow
   \textsc{ComputeWristPosition}(\mathbf{c}_h,\mathbf{R}_w)$\;
$\mathbf{x}_{\mathrm{pre}} \leftarrow
   \mathbf{x}_w + \delta\, \mathbf{n}_h$\;
$\mathbf{q}_{\mathrm{open}}  \leftarrow \textsc{OpenHandPose}()$\;
$\mathbf{q}_{\mathrm{pre}}   \leftarrow \textsc{PreShapeHandPose}()$\;
$\mathbf{q}_{\mathrm{grasp}} \leftarrow
   \textsc{ForceClosurePose}(d_h)$\;
$\tau \leftarrow \emptyset$\;

\BlankLine
\tcc{Approach: wrist moves from pre-grasp to grasp; fingers shape up.}
$\tau \leftarrow \tau \cup
   \textsc{ExecutePhase}(\mathbf{x}_{\mathrm{pre}},\mathbf{x}_w,
   \mathbf{q}_{\mathrm{open}},\mathbf{q}_{\mathrm{pre}},
   T_{\mathrm{approach}})$\;

\BlankLine
\tcc{Grasp: wrist fixed; fingers close to force closure.}
$\tau \leftarrow \tau \cup
   \textsc{ExecutePhase}(\mathbf{x}_w,\mathbf{x}_w,
   \mathbf{q}_{\mathrm{pre}},\mathbf{q}_{\mathrm{grasp}},
   T_{\mathrm{grasp}})$\;

\BlankLine
\tcc{Drag: follow target joint axis through contact.}
\For{$t = 1$ \KwTo $T_{\mathrm{drag}}$}{
   $\alpha \leftarrow t / T_{\mathrm{drag}}$\;
   \eIf{$m$ is revolute}{
       $(\mathbf{c}_t,\mathbf{R}_t) \leftarrow
        \textsc{RotateAroundJoint}(
         \mathbf{c}_h,\mathbf{R}_w,\mathcal{M},\alpha)$\;
   }{
       $(\mathbf{c}_t,\mathbf{R}_t) \leftarrow
        \textsc{TranslateAlongNormal}(
         \mathbf{c}_h,\mathbf{R}_w,\mathbf{n}_h,\alpha)$\;
   }
   $\mathbf{x}_t \leftarrow
      \textsc{ComputeWristPosition}(\mathbf{c}_t,\mathbf{R}_t)$\;
   $\tau \leftarrow \tau \cup
      \textsc{StepAndRecord}(\mathbf{x}_t,\mathbf{R}_t,
      \mathbf{q}_{\mathrm{grasp}})$\;
}

\BlankLine
\tcc{Release: open fingers and retract wrist along outward normal.}
$\tau \leftarrow \tau \cup
   \textsc{ReleaseAndRetract}(
   \mathbf{c}_{\mathrm{final}},\mathbf{R}_{\mathrm{final}},
   \mathbf{q}_{\mathrm{grasp}},\mathbf{q}_{\mathrm{open}},
   T_{\mathrm{release}})$\;

\BlankLine
Persist $\tau$ as the reference contact trajectory used by the
DragMesh~2 environment (initial state writing), the trajectory
tracking baseline, and the motion-scale reference of the evaluation
protocol\;
\Return $\tau$\;
\end{algorithm}

\section{Additional Experimental Results}
\label{sec:exp}

This appendix reports diagnostic results that are omitted from the
compact main text. Section~\ref{sec:exp_main} gives the main success
table; the appendix focuses on additional qualitative visualizations,
rollout-level behavior, strong-damping diagnostics, extended
fine-tuning, damping-range expansion, and ablations.
All multi-episode cells use the expert-grasp initialization and 20
episodes per deterministic or stochastic execution mode unless stated
otherwise. These diagnostic studies are auxiliary to the seven-object
main comparison; they explain failure modes and checkpoint-selection
effects, but they do not replace the aggregate success and ablation
tables in Section~\ref{sec:exp_main}.

In the checkpoint-diagnostic subsections, the \emph{base policy}
denotes a PICA reference policy trained with the full physical-signal
mechanism on the diagnostic object. \emph{Base ($N$\,ep)} denotes its
checkpoint after $N$ epochs. The contact-stabilization modules ARAM
and Reconfig (Appendix~\ref{sec:reward}) are applied on top of the
base policy for a further 50 epochs, giving the \emph{ARAM},
\emph{Reconfig}, and \emph{Both} fine-tunes. \textsc{Selected} and
\textsc{Overtrained} are used only for single-rollout visualization
diagnostics: \textsc{Selected} is base (150\,ep) followed by the Both
fine-tune, while \textsc{Overtrained} continues that fine-tune well
beyond it.

\paragraph{Training-curve scope.}
We omit raw training reward curves from the appendix, because reward
scales are not identical once contact regularizers, auxiliary terms,
and fine-tuning modules are introduced. The baseline evidence should
therefore remain evaluation-side success, progress, \texttt{clip099},
and \texttt{detach\_proxy} across damping conditions. If additional
curves are needed for diagnostics, they should compare evaluation
metrics rather than raw rewards; only within the same training recipe
is a reward-versus-OOD-success curve directly interpretable.

\subsection{Qualitative Simulation and Hardware Visualizations}
\label{sec:qual_app}

\begin{figure*}[t]
\centering
\setlength{\tabcolsep}{2pt}
\renewcommand{\arraystretch}{0.56}
\begin{tabular}{@{}c ccc@{}}
 & \footnotesize Approach & \footnotesize Grasp & \footnotesize Drag \\
\makecell[r]{\scriptsize 46440\\[-0.2em]\scriptsize StorageFurn.\\[-0.2em]\scriptsize drawer} &
\includegraphics[width=0.255\textwidth]{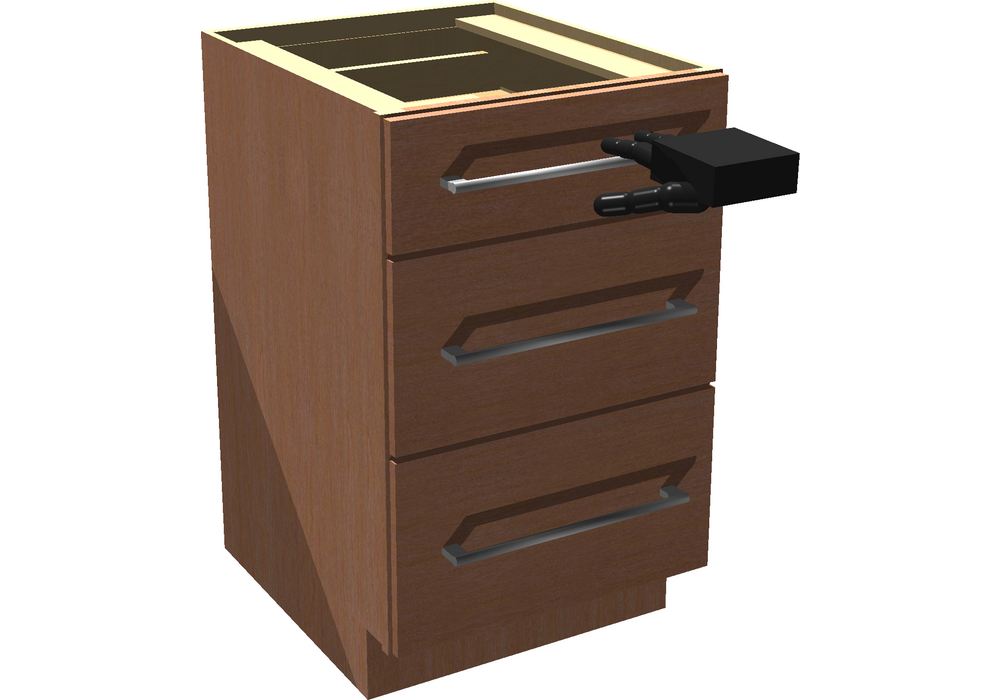} &
\includegraphics[width=0.255\textwidth]{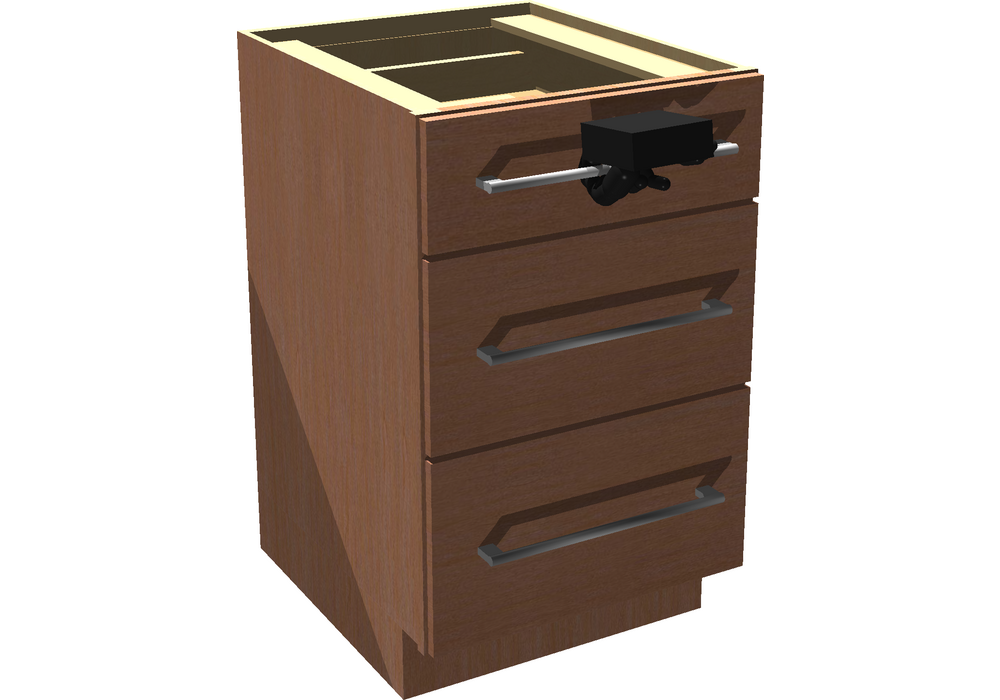} &
\includegraphics[width=0.255\textwidth]{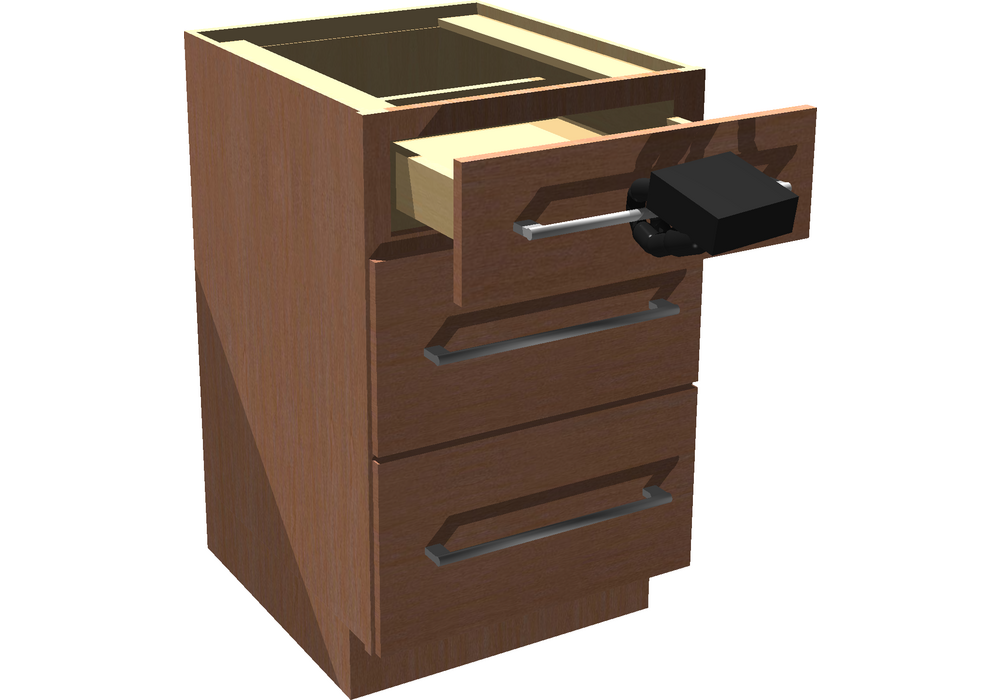} \\
\makecell[r]{\scriptsize 12583\\[-0.2em]\scriptsize Dishwasher\\[-0.2em]\scriptsize door} &
\includegraphics[width=0.255\textwidth]{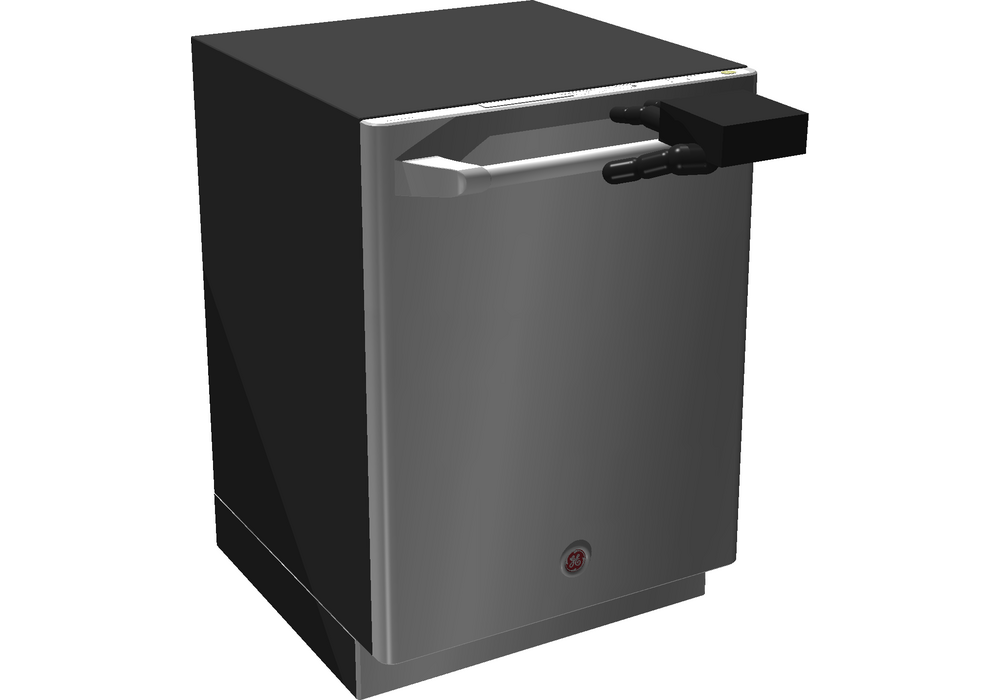} &
\includegraphics[width=0.255\textwidth]{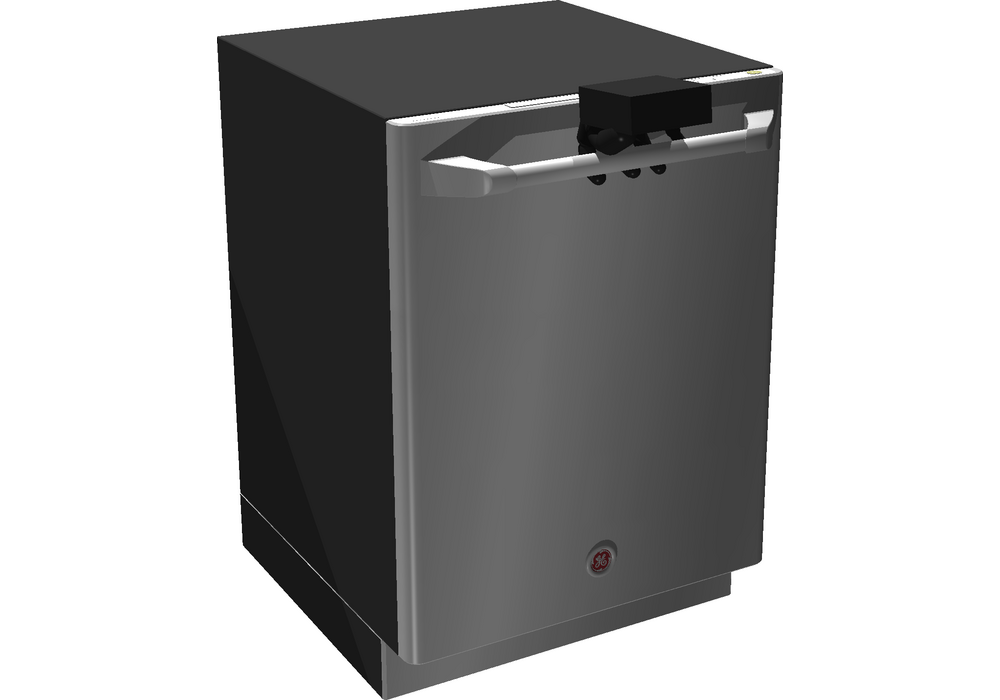} &
\includegraphics[width=0.255\textwidth]{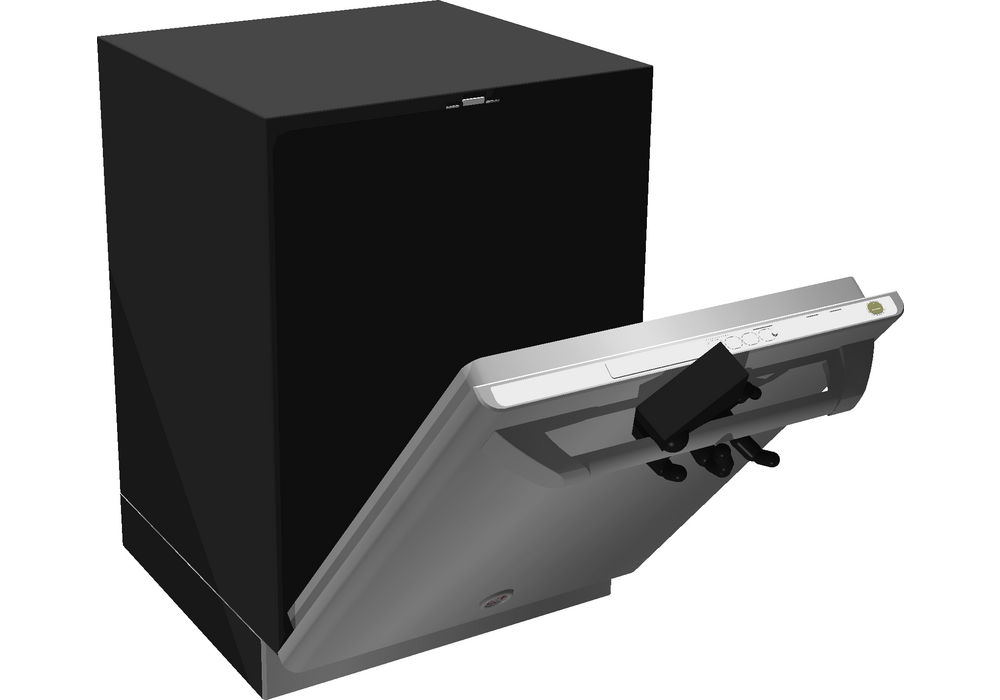} \\
\makecell[r]{\scriptsize 7310\\[-0.2em]\scriptsize Microwave\\[-0.2em]\scriptsize door} &
\includegraphics[width=0.255\textwidth]{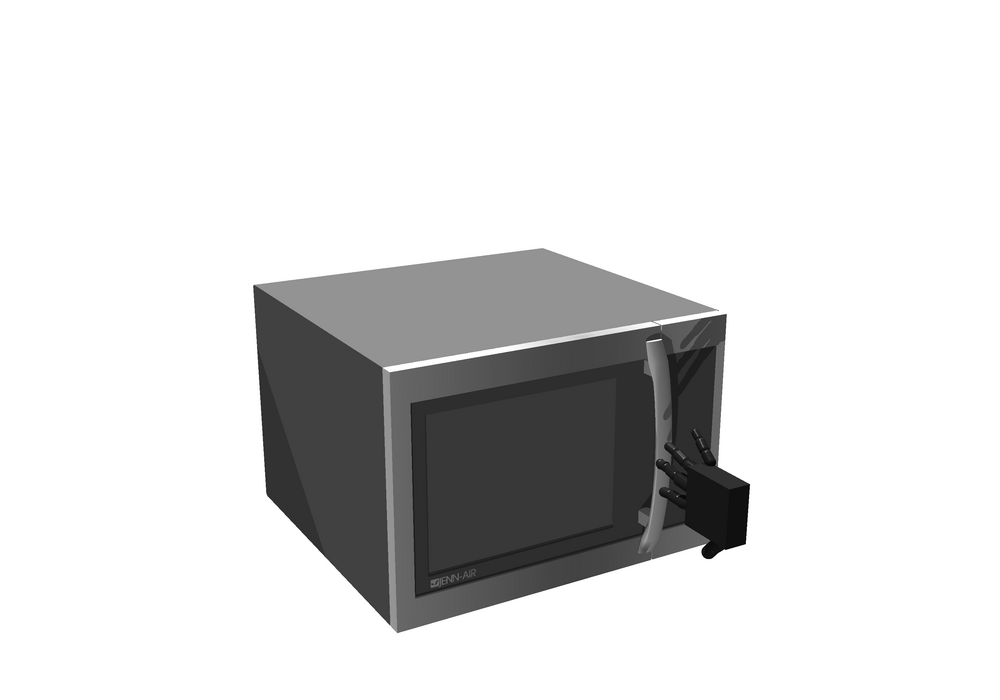} &
\includegraphics[width=0.255\textwidth]{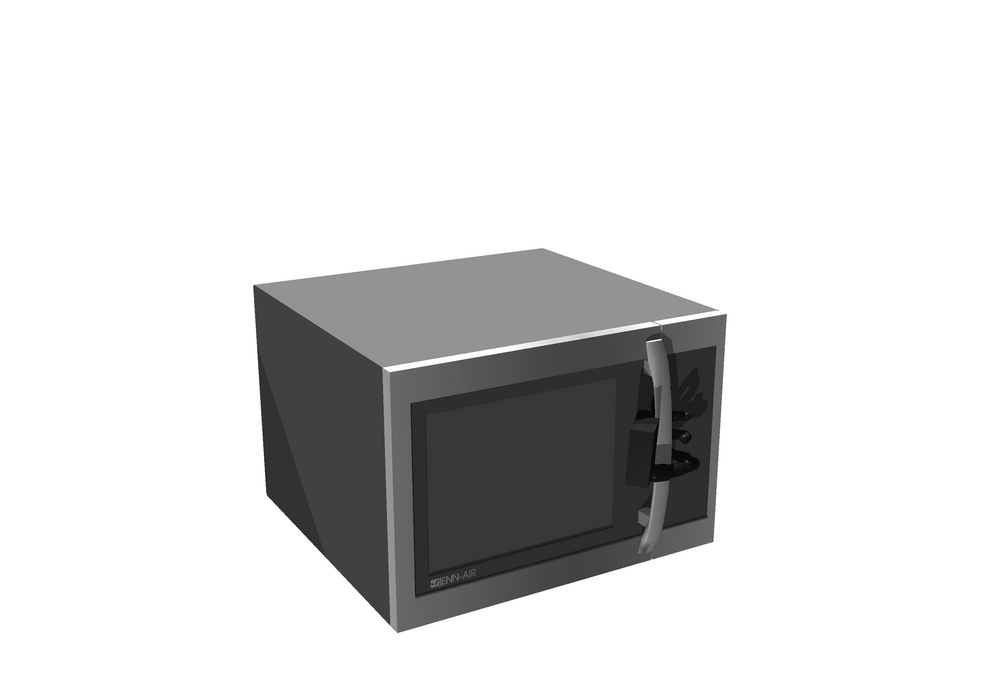} &
\includegraphics[width=0.255\textwidth]{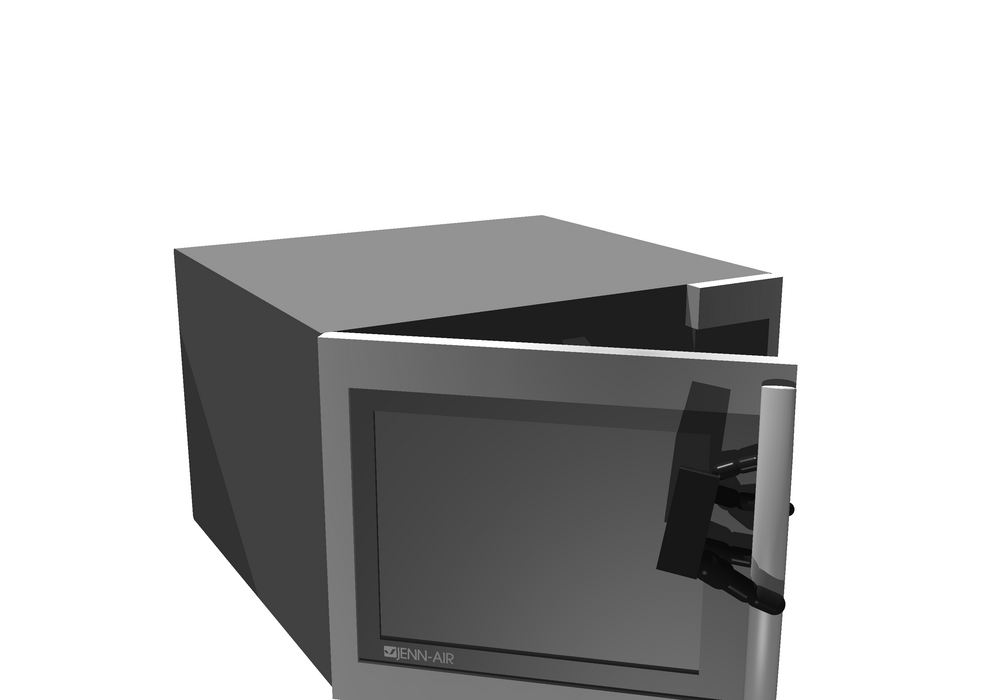} \\
\end{tabular}
\caption{Additional simulated approach, grasp, and drag snapshots on
Three generated object trajectories.}
\label{fig:app_sim_stage}
\end{figure*}

\begin{figure*}[t]
\centering
\setlength{\tabcolsep}{0pt}
\renewcommand{\arraystretch}{0.68}
\begin{tabular}{@{}ccccc@{}}
\includegraphics[width=0.195\textwidth,trim=900 180 850 600,clip]{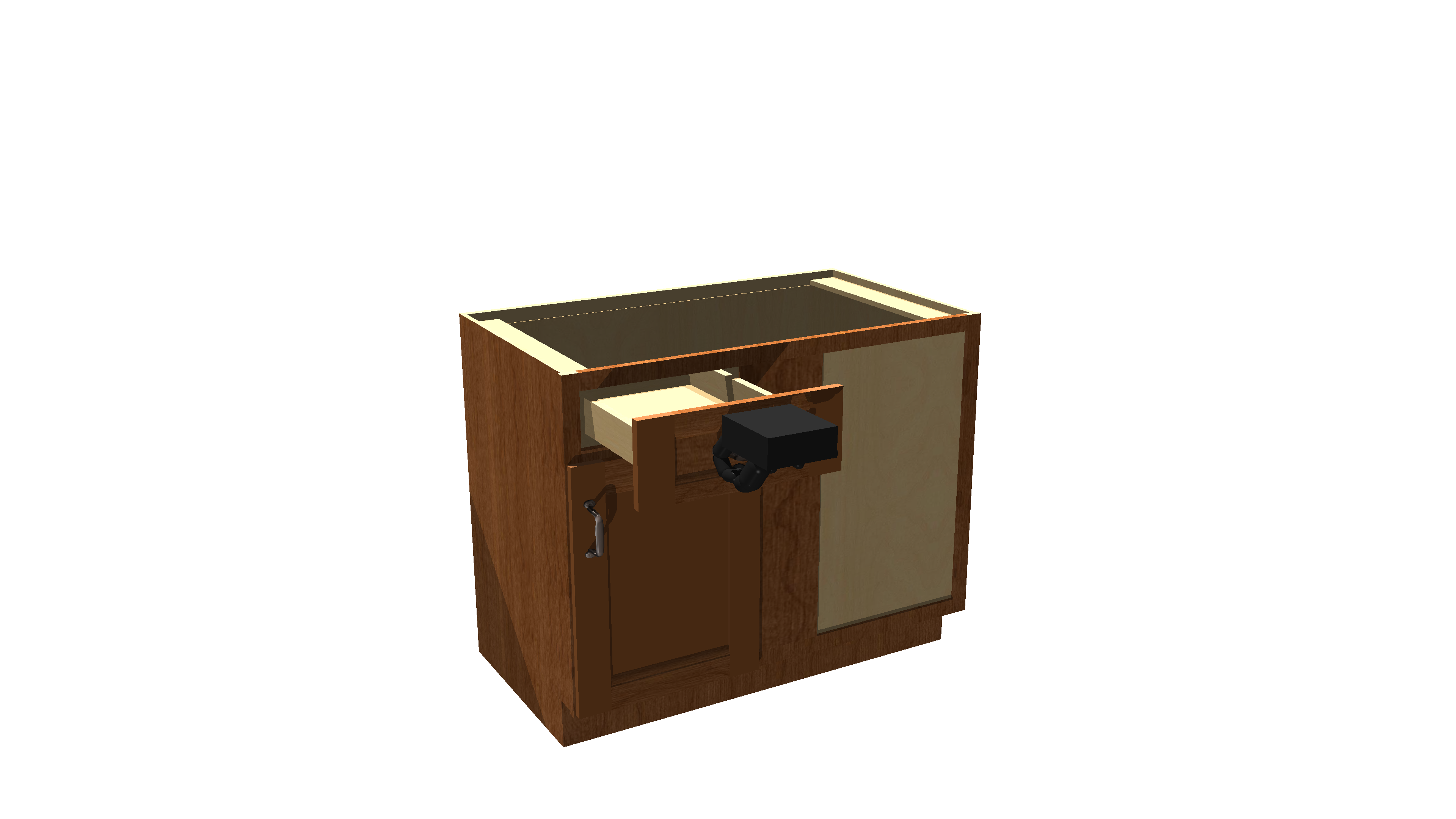} &
\includegraphics[width=0.195\textwidth,trim=1050 0 650 550,clip]{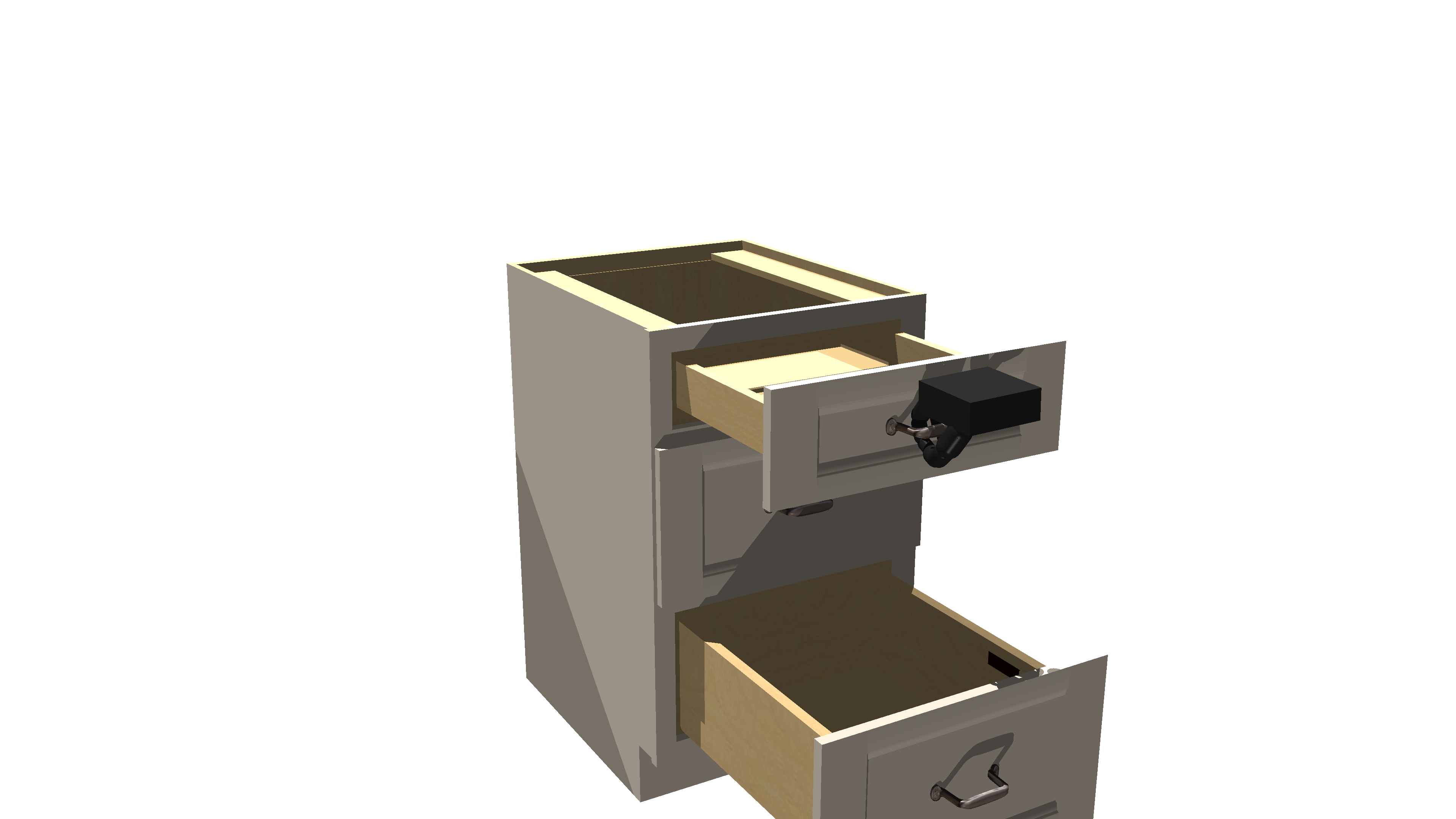} &
\includegraphics[width=0.195\textwidth,trim=900 180 850 600,clip]{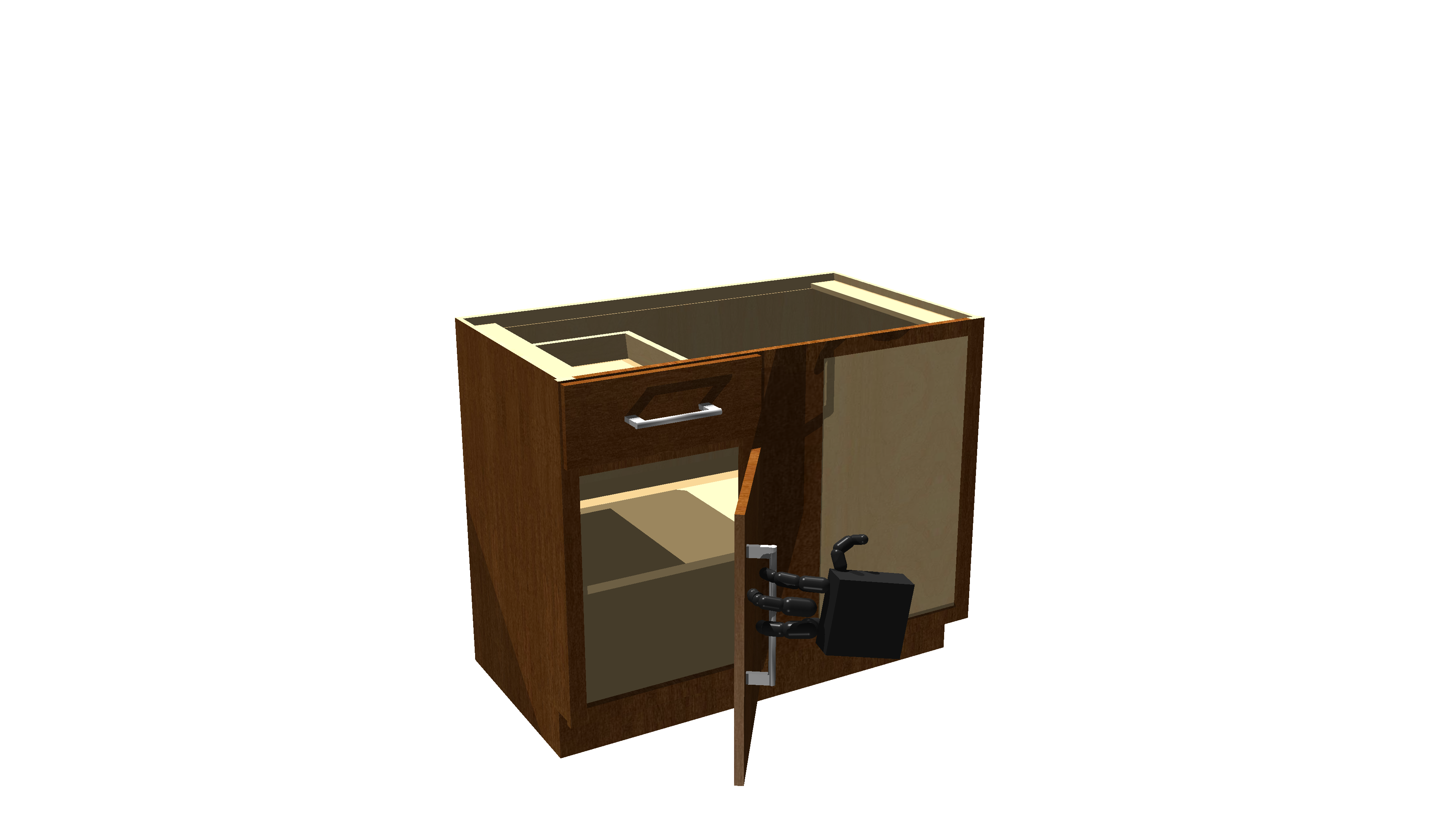} &
\includegraphics[width=0.195\textwidth,trim=900 100 650 600,clip]{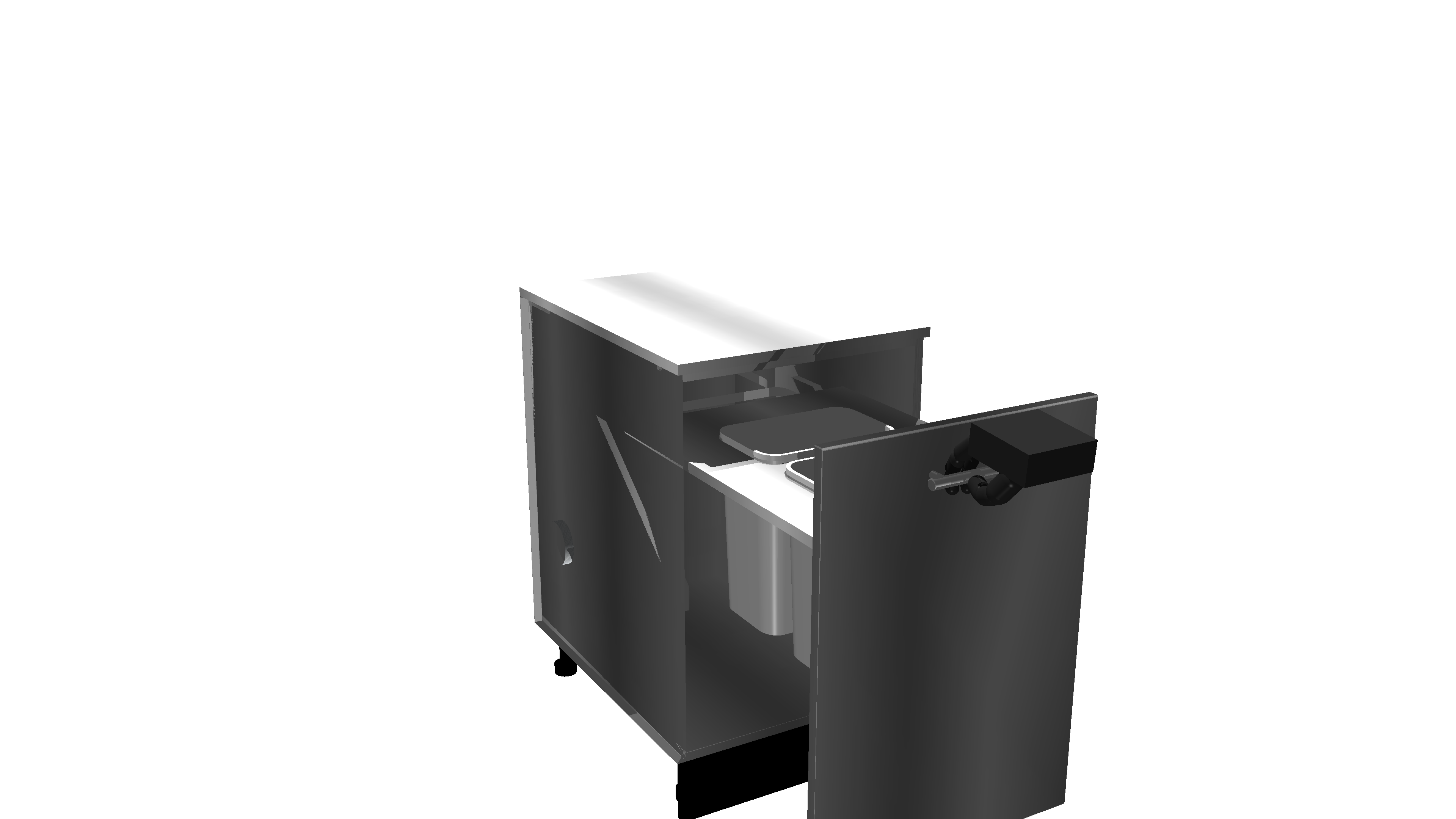} &
\includegraphics[width=0.195\textwidth,trim=1100 0 700 600,clip]{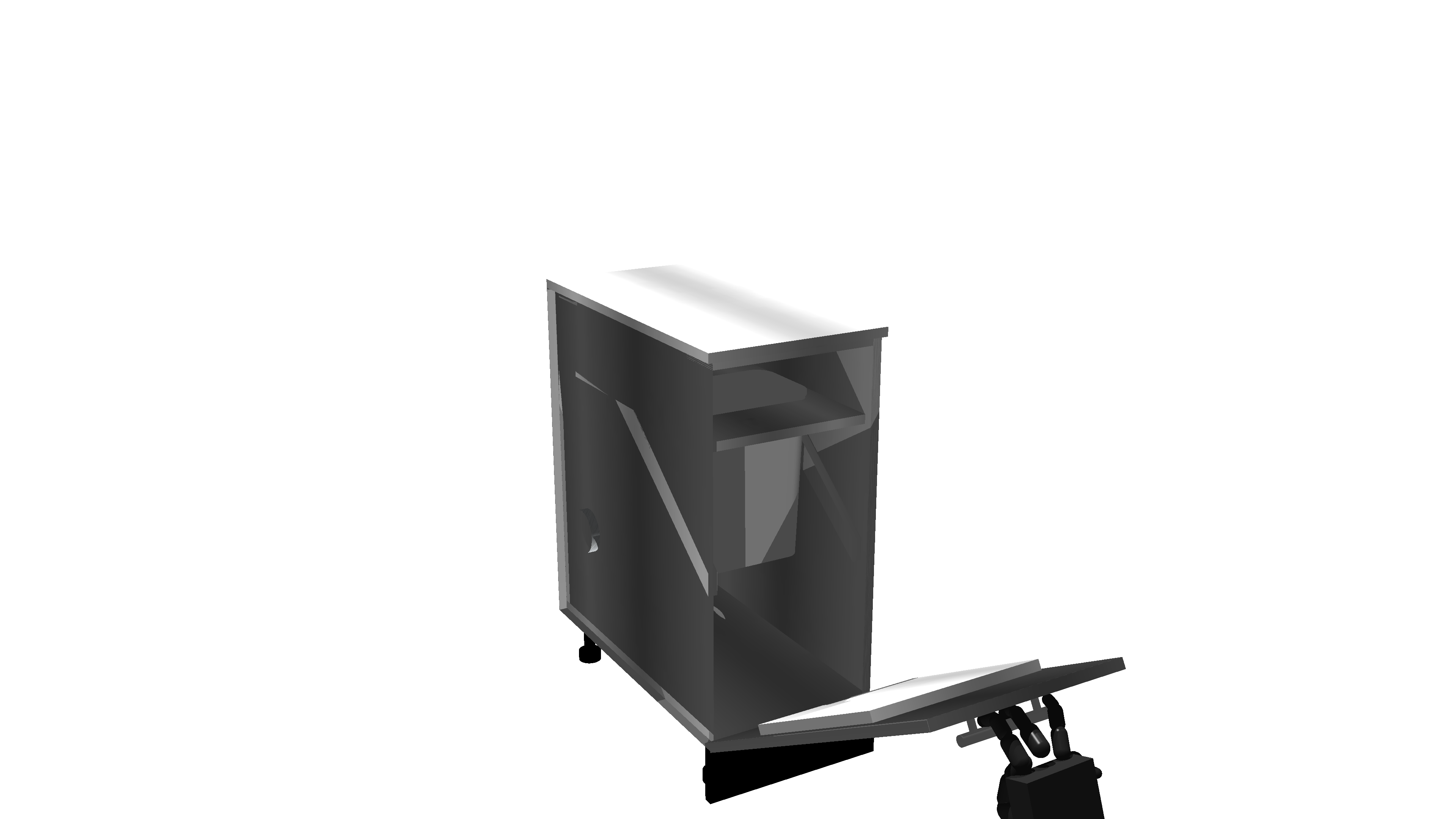} \\
\makecell{\scriptsize 40147\\[-0.2em]\scriptsize StorageFurn.\\[-0.2em]\scriptsize slider drawer} &
\makecell{\scriptsize 44962\\[-0.2em]\scriptsize StorageFurn.\\[-0.2em]\scriptsize slider drawer} &
\makecell{\scriptsize 48513\\[-0.2em]\scriptsize StorageFurn.\\[-0.2em]\scriptsize hinge door} &
\makecell{\scriptsize 102996\\[-0.2em]\scriptsize TrashCan\\[-0.2em]\scriptsize slider drawer} &
\makecell{\scriptsize 103008\\[-0.2em]\scriptsize TrashCan\\[-0.2em]\scriptsize hinge door} \\
\end{tabular}
\caption{Additional terminal-stage renders showing category and
joint-type diversity in the generated trajectory dataset.}
\label{fig:app_sim_diversity}
\end{figure*}

\begin{figure*}[t]
\centering
\setlength{\tabcolsep}{2pt}
\renewcommand{\arraystretch}{0.5}
\begin{tabular}{@{}ccc@{}}
\footnotesize Approach & \footnotesize Grasp & \footnotesize Open \\
\includegraphics[width=0.31\textwidth]{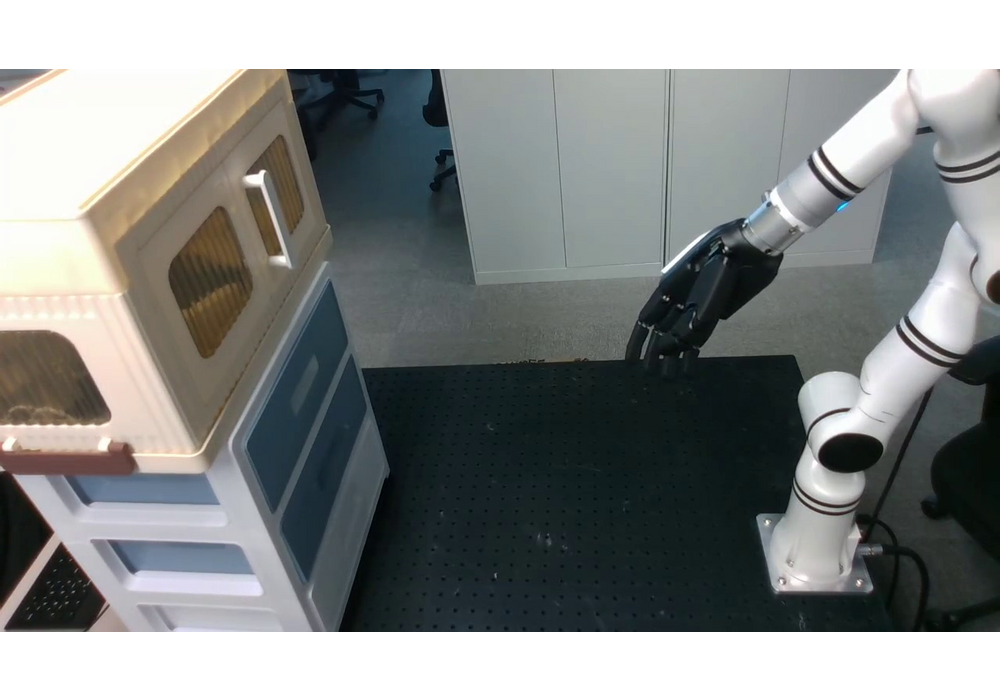} &
\includegraphics[width=0.31\textwidth]{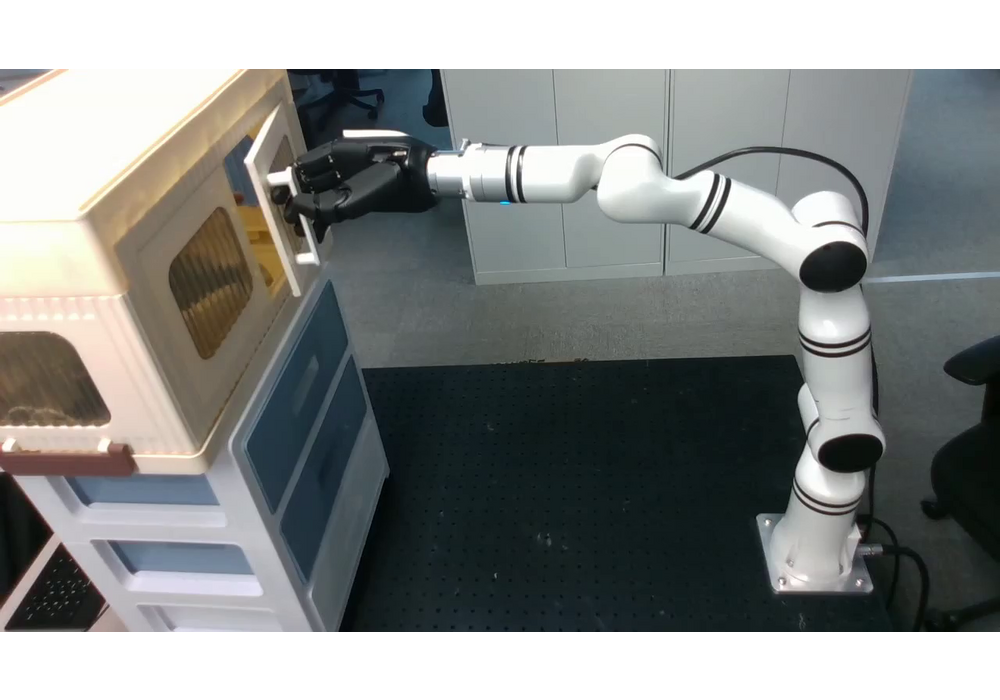} &
\includegraphics[width=0.31\textwidth]{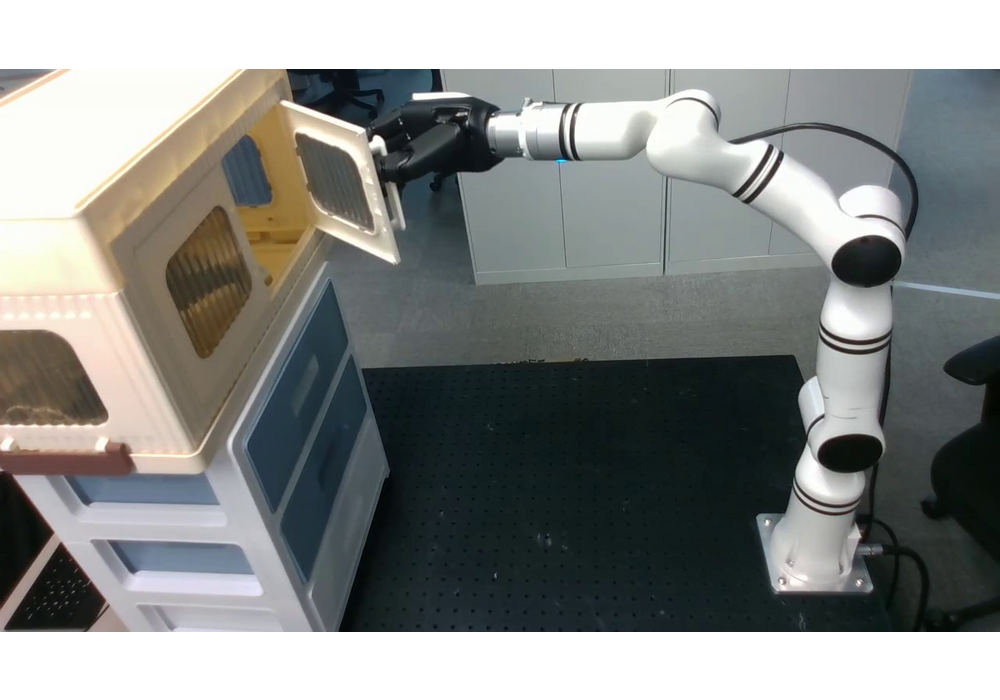} \\
\end{tabular}
\caption{Hardware stage snapshots extracted from the supplementary
stage video.}
\label{fig:app_real_stage}
\end{figure*}

Figures~\ref{fig:app_sim_stage} and~\ref{fig:app_real_stage} provide
additional visual evidence using frames not shown in the main text.
The simulation snapshots illustrate that the same contact-driven
formulation covers both prismatic drawers and revolute doors.
Figure~\ref{fig:app_sim_stage} shows approach, grasp, and drag stages
for three object instances, and
Figure~\ref{fig:app_sim_diversity} shows additional terminal-stage
examples. For Figure~\ref{fig:app_sim_diversity}, each render replays
the stored right-hand trajectory state using the simulator's 51-DoF
floating SMPL-X right-hand model: 3 virtual wrist-translation DoFs,
3 wrist-rotation DoFs, and 45 finger DoFs. The hardware frames are extracted from the
supplementary stage video and are included as a qualitative
feasibility check. They are not pooled with the 20-episode simulation
statistics and do not constitute a separate real-world quantitative
benchmark.

\subsection{Rollout-Level Diagnostics}
\label{sec:rollout}

To interpret the contact behavior behind the numerical metrics, we
compare single-rollout visualizations for two checkpoints:
\textsc{Selected} denotes the base policy (150\,ep) followed by a
50-epoch Both-module fine-tune, and \textsc{Overtrained} denotes a
checkpoint obtained by continuing that fine-tune well beyond it. Only one rollout per object, execution mode, and damping
setting is saved. Thus, this section describes behavior patterns and
is not pooled into the multi-episode success statistics.

\begin{table}[!htbp]
\centering
\caption{Single-rollout diagnostics for representative checkpoints.}
\label{tab:rollout}
\small
\begin{tabular}{llcccccc}
\toprule
Checkpoint & Object & Success & Mean Prog. & Best Prog. & Steps & $L_2$ \\
\midrule
\textsc{Selected} & 7310  & 5/6 & 0.451 & 0.564 & 33.3 & 3.624 \\
\textsc{Selected} & 45936 & 3/6 & 0.285 & 0.597 & 66.0 & 4.132 \\
\textsc{Overtrained} & 7310  & 0/6 & 0.175 & 0.288 &  6.0 & 3.351 \\
\textsc{Overtrained} & 45936 & 2/6 & 0.199 & 0.548 & 36.8 & 3.104 \\
\bottomrule
\end{tabular}
\end{table}

The video evidence is consistent with the numerical summary. On
7310, \textsc{Selected} succeeds in 5 of 6 rollouts and, even when it
fails under strong damping, maintains a relatively continuous
pulling pose. The failure mode resembles a correctly aligned pulling
direction with insufficient contact output to overcome the elevated
damping load. The trend on 45936 is similar: \textsc{Selected}
completes some rollouts under nominal and mid damping, and under
$\times4$ it fails with relatively long episode durations, indicating
that the policy remains in contact interaction rather than detaching
immediately.

The \textsc{Overtrained} checkpoint shows a different pattern. It fails
on every rollout of 7310, with a mean episode length of only $6.0$
steps and a best progress of $0.288$. It does not apply small and
steady pulling actions, but rapidly enters a failure or abnormal
contact state. The video shows that \textsc{Overtrained} develops
visible perturbations after failing to pull the part open, and these
perturbations do not translate into target-joint progress. This
behavior is consistent with the low mean action $L_2$, short episode
length, and low progress. On 45936, \textsc{Overtrained} succeeds only
under nominal damping and degrades sharply at higher damping,
indicating that subsequent fine-tuning does not improve robustness
to contact load and may instead weaken the stable temporal pulling
strategy.

This visualization diagnostic supports a finer claim than success
alone. The dominant failure mode of \textsc{Selected} is a largely
correct contact direction and pulling intent with insufficient
sustained output, whereas \textsc{Overtrained} fails to maintain stable
contact response under resistance and switches to ineffective
perturbation. These visualizations are used only to interpret the
diagnostic logs; the main paper's quantitative claims remain the
seven-object simulation results in Section~\ref{sec:exp_main}.

\subsection{OOD Damping Robustness}
\label{sec:ood}

The $\times4$ damping setting is analyzed separately on diagnostic
object 45936 because it most directly exposes stability under strong load.
Under deterministic
execution, the base policy (150\,ep), the base policy (200\,ep),
the Reconfig fine-tune, and the Both-module fine-tune all retain
$\times4$ success in the $0.50$--$0.55$ range, while
the base policy (300\,ep) and the base policy (500\,ep) drop to
roughly $0.10$ with \texttt{clip099} approaching $0.99$. This
indicates that the base policy undergoes a relatively rapid OOD $\times4$
collapse between epochs 200 and 300 and then settles into a
high-saturation, low-robustness regime. Longer training is therefore
not sufficient for physical robustness and can push the policy
toward a highly saturated nominal-dynamics solution. Under
stochastic execution, the Both-module fine-tune is the only variant in
the available logs to reach $0.55$ at $\times4$, suggesting that the
combined fine-tuning is more stable when dynamics perturbation and
sampling noise act together.

\begin{table}[t]
\centering
\caption{Object-45936 per-checkpoint diagnostics at $\times4$ damping.
Det and Stoch rows are reported separately.}
\label{tab:ood}
\small
\begin{tabular}{llccccc}
\toprule
Variant & Mode & Success $\times4$ & Prog. $\times4$ & Return $\times4$
        & clip099 $\times4$ & detach $\times4$ \\
\midrule
base (150 ep)      & Det & 0.55 & 0.398 & 138.0 & 0.896 & 0.45 \\
base (200 ep)      & Det & 0.50 & 0.425 & 137.3 & 0.972 & 0.50 \\
base (300 ep)      & Det & 0.10 & 0.206 & 19.5  & 0.986 & 0.90 \\
base (500 ep)      & Det & 0.10 & 0.236 & 27.8  & 0.993 & 0.90 \\
ARAM (50 ep)      & Det & 0.25 & 0.382 & 88.6  & 0.969 & 0.75 \\
Reconfig (50 ep)  & Det & 0.50 & 0.439 & 141.1 & 0.981 & 0.50 \\
Both (50 ep)      & Det & 0.55 & 0.449 & 151.1 & 0.971 & 0.45 \\
\midrule
base (150 ep)      & Stoch & 0.30 & 0.334 & 83.7 & 0.936 & 0.70 \\
base (200 ep)      & Stoch & 0.10 & 0.249 & 31.3 & 0.967 & 0.90 \\
base (300 ep)      & Stoch & 0.15 & 0.318 & 56.5 & 0.939 & 0.85 \\
base (500 ep)      & Stoch & 0.30 & 0.201 & 48.5 & 0.926 & 0.70 \\
ARAM (50 ep)      & Stoch & 0.15 & 0.337 & 61.8 & 0.984 & 0.85 \\
Reconfig (50 ep)  & Stoch & 0.30 & 0.383 & 96.4 & 0.986 & 0.70 \\
Both (50 ep)      & Stoch & 0.55 & 0.420 & 143.5 & 0.973 & 0.45 \\
\bottomrule
\end{tabular}
\end{table}

\subsection{Limits of Extended Fine-Tuning}
\label{sec:contft}

The OOD-robustness collapse in the base policy occurs between epochs
200 and 300. A natural question is whether the 50 epochs of
Both-module fine-tuning fall in an early-stopping window or
whether the Both-module fine-tune is still under-trained. Starting from
the Both-module fine-tune (total epoch 200) and keeping the
training recipe fixed, we continue training for another 200 epochs to
total epoch 400 and reevaluate checkpoints at epoch 250,
epoch 300, and epoch 399 on object 45936. This diagnostic uses a
separate controlled re-evaluation on object 45936 with 20 episodes per
cell and is intended only to analyze within-run fine-tuning trends
rather than replace the seven-object main comparison.

\begin{table}[t]
\centering
\caption{Extending fine-tuning beyond the Both fine-tune. The
training-reward-best checkpoint at epoch 399 still lies in the
degraded regime, while \texttt{clip099} drifts monotonically upward.}
\label{tab:contft}
\scriptsize
\setlength{\tabcolsep}{3pt}
\begin{tabular}{lccccccc}
\toprule
Checkpoint & Det $\times1$ & Det $\times2$ & Det $\times4$
           & Stoch $\times1$ & Stoch $\times2$ & Stoch $\times4$ & clip099 $\times4$ \\
\midrule
Both fine-tune (200 ep)              & 1.00 & 0.85 & 0.00 & 0.95 & 0.70 & 0.20 & - \\
continued (250 ep)             & 1.00 & 0.25 & 0.10 & 0.90 & 0.10 & 0.10 & 0.598 \\
continued (300 ep)            & 1.00 & 0.60 & 0.00 & 0.95 & 0.20 & 0.05 & 0.732 \\
continued (399 ep, reward-best)
                                & 0.90 & 0.30 & 0.00 & 0.95 & 0.15 & 0.10 & 0.869 \\
\bottomrule
\end{tabular}
\end{table}

From the training curves, extended training appears stable:
$\texttt{success\_mean}$ stays near $1.0$ throughout epochs
200--400, and $\texttt{reward\_mean}$ rises slowly from roughly 255
to 260. The OOD evaluation shows the opposite trend. Deterministic
$\times2$ success drops immediately from $0.85$ for
the Both-module fine-tune to $0.25$ at epoch 250, partially
recovers to $0.60$ at epoch 300, and falls back to $0.30$ at
epoch 399. Stochastic $\times2$ likewise degrades from $0.70$ to
the $0.10$--$0.20$ range. Deterministic $\times4$ remains near the
$0.00$--$0.10$ floor, and extended training provides no stable
improvement. At the same time, \texttt{clip099} at $\times4$ grows
monotonically over epochs ($0.598 \to 0.732 \to 0.869$), indicating
slow saturation of the action distribution.

ARAM delays, but does not block, the drift toward saturated actions.
Across 200 additional epochs,
the Both-module fine-tune reproduces the OOD collapse that
the base policy undergoes between epochs 200 and 300, only stretched
to roughly four times the time scale. Even the checkpoint
automatically saved by the training framework as the best by training
reward (epoch 399) lies in the degraded regime, confirming
that checkpoint selection by training reward alone is insufficient
for OOD robustness. This supports OOD-based early stopping, rather
than monotone training reward, as a protocol-level principle for
checkpoint selection.

\subsection{Limits of Damping-Distribution Expansion}
\label{sec:damprangeexp}

The OOD $\times4$ collapse raises a second question: can fine-tuning
broaden the training damping range so that the policy encounters
strong-damping samples earlier and learns more stable contact
behavior? Starting from the Both-module fine-tune, we broaden the
damping scale interval from $[1.0, 2.0]$ to $[1.0, 4.0]$ while
keeping all reward terms, action-boundary regularizers, network
structure, and training hyperparameters fixed. We then fine-tune for
25 epochs on object 45936 to obtain
the broadened-damping fine-tune. The two rows are evaluated under the
same controlled setting on object 45936 and are used only to analyze
the effect of broadening the damping range.

\begin{table}[t]
\centering
\caption{Broadening the training damping range during fine-tuning.}
\label{tab:damprange}
\small
\begin{tabular}{lcccccc}
\toprule
Variant & Det $\times1$ & Det $\times2$ & Det $\times4$ & Stoch $\times1$ & Stoch $\times2$ & Stoch $\times4$ \\
\midrule
Both fine-tune (controlled re-eval) & 1.00 & 0.85 & 0.00 & 0.95 & 0.70 & 0.20 \\
Both + damping $[1,4]$ (25 ep)   & 1.00 & 0.50 & 0.05 & 0.95 & 0.35 & 0.00 \\
\bottomrule
\end{tabular}
\end{table}

Broadening the training damping range yields no stable improvement
at $\times4$: deterministic success rises only from $0.00$ to $0.05$,
while $\times2$ degrades noticeably (deterministic
$0.85 \to 0.50$, stochastic $0.70 \to 0.35$) and stochastic $\times4$
regresses overall. An additional check of intermediate checkpoints
shows that one intermediate epoch transiently reaches $0.25$
deterministic $\times4$ success, but its $\times2$ success drops in parallel
to roughly $0.55$, and the effect is not reproduced in later epochs.
Earlier exploration with the narrower range $[1.0, 2.5]$ combined
with delayed ARAM, and with the more aggressive range $[2.0, 4.0]$,
produces sharper degradation at $\times4$ and $\times2$ and is not listed in
the main table.

Broadening the training damping range alone provides limited benefit
under the current incremental-position control and reward-shaping
framework. The contact-interface capability required at $\times4$, namely
stable light pulling under sustained high load, may lie beyond what
the present 51-dimensional position-increment action channel and
force-free observation channel can represent. Further improvement in
OOD $\times4$ robustness therefore requires changes to the task or policy
interfaces, such as adding a wrist force or torque output dimension
on the action side, introducing contact force or torque feedback on
the observation side, or separating light pulling and heavy pulling
into different contact modes at the policy level via mode switching
or expert mixtures.

\subsection{Ablation Analysis}
\label{sec:ablation}

This diagnostic ablation analysis examines the effects of training
duration, damping randomization, and the two fine-tuning modules on
object 45936. Comparing
the base policy at 150, 200, 300, and 500 epochs shows that additional
training improves success under nominal and mid damping but does not
monotonically improve robustness under strong damping. The OOD $\times4$
degradation concentrates between epochs 200 and 300:
the base policy at 300\,ep already nearly matches the 500\,ep checkpoint on
$\times4$ success ($0.10$) and \texttt{clip099} ($0.986$ vs.~$0.993$).
The collapse is therefore not a slow over-saturation but a
relatively rapid behavioral transition. Continued fine-tuning shows
the same pattern under ARAM, stretched to roughly four times the
time scale. The direction does not reverse: between
epoch 200 and epoch 400, deterministic $\times2$ success
degrades from $0.85$ to the $0.30$ range while \texttt{clip099}
rises monotonically. These results support the observation that
optimizing only the task reward selects high-saturation,
low-robustness contact policies, so the protocol must report action
saturation and detachment alongside success and use OOD-based
criteria rather than the training reward for checkpoint selection.

Broadening the training damping distribution alone does not resolve
the OOD $\times4$ failure. The extended-damping fine-tuning variant
improves deterministic $\times4$ only from $0.00$ to $0.05$ and
introduces significant degradation at $\times2$. Section~\ref{sec:damprangeexp}
gives the detailed analysis. Within the incremental-position control
and reward-shaping framework, adjusting the training damping
interval alone provides limited benefit. Further improvement at
$\times4$ requires new mechanisms at the action channel, contact
observation, or policy level rather than only a different training
distribution.

The three fine-tune variants indicate different roles for the physical
modules. The ARAM fine-tune improves deterministic nominal-damping
success but yields no robustness gain at $\times4$, suggesting that
constraining high-effort actions alone does not solve contact
recovery under strong damping. The Reconfig fine-tune raises
deterministic $\times2$ success to $0.95$ and preserves higher $\times4$
progress, suggesting that contact reconfiguration helps recovery
under moderate resistance. The Both-module fine-tune reaches $0.55$
success and $0.420$ progress under stochastic $\times4$ while keeping
\texttt{detach\_proxy} at $0.45$, the most stable stochastic
strong-damping result in the available logs. The two fine-tuning
modules are therefore not simple substitutes: ARAM acts more
directly on high-effort action behavior, while Reconfig is closer to
contact recovery. Their combination is most useful when dynamics
perturbation and sampling noise occur together.

\subsection{Summary of Diagnostic Findings}
\label{sec:findings}

The diagnostics support a narrower interpretation of the appendix
experiments. PICA's contact regularizers, dynamics randomization, and
temporal auxiliary supervision should be read as a coupled protocol:
GLA provides temporal capacity, but stable contact response requires
pairing that capacity with explicit contact-maintenance and
action-regularization signals. In the object-45936 diagnostic logs,
the combined ARAM/Reconfig fine-tune gives the strongest stochastic
$\times4$ result among the diagnostic variants considered here, while
temporal encoding alone does not prevent saturation and detachment
shortcuts.

Across extended fine-tuning and damping-range expansion, task reward
and nominal success are not reliable checkpoint selectors for
strong-load robustness. Longer training can preserve nominal
performance while weakening $\times2/\times4$ robustness, and
broadening the damping range alone gives limited $\times4$ benefit
while degrading mid-damping behavior. These trends motivate reporting
success together with \texttt{clip099}, \texttt{detach\_proxy},
progress, and damping-conditioned performance, rather than treating a
single nominal score as sufficient.

\subsection{Relationship Between Physical Diagnostics and Temporal
Encoding}
\label{sec:phys-temp}

The central claim that follows from these empirical results is that,
for contact-driven articulated-object manipulation, physical
diagnostics, the robustness summary, and temporal encoding must be
designed jointly into the protocol. Temporal encoding alone can let
the policy reach a nominal-dynamics shortcut faster; physical shaping
alone attains some robustness but lacks fine-grained modeling of
historical contact response. The policy is more likely to extract a
representation consistent with contact state only when the task
protocol explicitly specifies contact-maintaining behavior
and auxiliary supervision guides the temporal encoder to predict
observable responses under such behavior. The experiments above
provide progressive evidence for this position.

\subsection{Limitations}
\label{sec:limits}

Two caveats apply to the numbers reported here. First, each cell
aggregates 20 episodes, so the standard error of a success estimate is
approximately $0.10$; small differences under strong damping, such as
$0.15$ versus $0.30$ at $\times4$, should be read as the same order
rather than as strictly ranked, and our conclusions rely on aggregate
trends across objects, damping multipliers, and execution modes rather
than on any single cell.

Second, because the observation channel provides no force or tactile
signal, the policy infers contact state only indirectly from
kinematic error. Under strong damping, this is the main caveat behind
the residual light-pulling failures in the diagnostic results.

\subsection{Future Work}
\label{sec:additional_future_work}

Future work can extend the action and observation interfaces toward
force-aware control and richer contact-response supervision. Within
the strong-load regime, useful directions include separating light and
heavy pulling into distinct contact modes and expanding the
auxiliary-supervision targets to quantities such as contact normals or
sliding velocity. Together with the whole-body loco-manipulation
extension outlined in the main text, these directions define a path
toward contact-rich articulated manipulation that remains
contact-stable under wider dynamics.

\section{Implementation Details and Inference Settings}
\label{app:impl}

This appendix collects the implementation parameters omitted from
the main text. These values support reproducibility and are not
methodological contributions; they are placed here so the main text
retains the task definition, evaluation protocol, and the key
diagnostic conclusions.

\subsection{Task and Control Parameters}
\label{app:taskhp}

Table~\ref{tab:taskhp} lists the task and control constants shared by
all policies: the success-threshold fraction, action scaling, episode
length, and the temporal-encoder and network dimensions.

\begin{table}[ht]
\centering
\caption{Task, control, and network parameters.}
\label{tab:taskhp}
\small
\begin{tabular}{lcl}
\toprule
Symbol & Value & Meaning \\
\midrule
$\rho$ & 0.5 & success-threshold fraction on the reference motion range \\
$\alpha$ & 0.05 & action-to-PD-target increment scale \\
$T_{\max}$ & 300 & maximum episode length \\
$L$ & 16 & contact-history window length \\
$K$ & 5 & causal auxiliary prediction window length \\
history token dim & 102 & dim of $[\text{tracking error}, \text{previous action}]$ \\
GLA heads & 4 & attention heads in the GLA encoder \\
token embedding dim & 128 & history-token projection dimension \\
MLP hidden dims & $[512,512,256]$ & current-state encoder hidden sizes \\
activation & ELU & MLP activation \\
\bottomrule
\end{tabular}
\end{table}

\subsection{Reward and Termination Parameters}
\label{app:rewhp}

Table~\ref{tab:rewhp} lists the reward weights, contact thresholds, and
one-shot bonuses and penalties used by the reference policy.

\begin{table}[ht]
\centering
\caption{Reward and termination parameters.}
\label{tab:rewhp}
\small
\begin{tabular}{lcl}
\toprule
Symbol & Value & Meaning \\
\midrule
$w_{\mathrm{dist}}$    & 0.25      & linear penalty on palm--handle distance \\
$w_{\mathrm{near}}$    & 0.10      & near-distance contact-keep bonus \\
$\kappa$               & 5.0       & decay coefficient of the near-distance bonus \\
$w_{\mathrm{task}}$    & 250       & weight on target-joint progress \\
$w_{\mathrm{act}}$     & 0.002     & weight on action energy penalty \\
$w_{\mathrm{time}}$    & 0.05      & per-step time cost \\
$d_{\mathrm{detach}}$  & 0.10\,m   & detachment-failure threshold \\
$r_{\mathrm{detach}}$  & $-50$     & one-shot detachment penalty \\
$r_{\mathrm{success}}$ & $+100$    & one-shot success reward \\
$a_{\mathrm{sat}}$     & 0.90      & soft action-saturation threshold \\
$w_{\mathrm{bound}}$   & 20        & weight on action-boundary regularizer \\
$d_{\mathrm{safe}}$    & 0.08\,m   & soft contact-distance threshold \\
$w_{\mathrm{contact}}$ & 0.5       & weight on contact-distance regularizer \\
\bottomrule
\end{tabular}
\end{table}

\subsection{Auxiliary-Supervision Parameters}
\label{app:auxhp}

Table~\ref{tab:auxhp} lists the channel weights of the causal-window
auxiliary loss and the warmup schedule of its overall weight.

\begin{table}[ht]
\centering
\caption{Auxiliary-supervision parameters.}
\label{tab:auxhp}
\small
\begin{tabular}{lcl}
\toprule
Symbol & Value & Meaning \\
\midrule
$\omega_q$         & 1.0 & weight on recent object joint response loss \\
$\omega_d$         & 1.0 & weight on max palm--handle distance loss \\
$\omega_b$         & 0.5 & weight on detachment-risk proxy loss \\
$\omega_e$         & 0.5 & weight on max tracking stress loss \\
$w_{\mathrm{aux}}$ & $0\rightarrow 0.01$ & linear warmup range \\
\bottomrule
\end{tabular}
\end{table}

\subsection{PPO Training Parameters}
\label{app:ppohp}

Table~\ref{tab:ppohp} lists the PPO optimization hyperparameters; they
follow standard on-policy settings and are not tuned per object.

\begin{table}[ht]
\centering
\caption{PPO training hyperparameters.}
\label{tab:ppohp}
\small
\begin{tabular}{lc}
\toprule
Hyperparameter & Value \\
\midrule
parallel environments     & 64 \\
rollout length            & 32 \\
discount $\gamma$         & 0.99 \\
GAE $\lambda$             & 0.95 \\
learning rate             & $3\times 10^{-4}$ \\
PPO clip                  & 0.2 \\
minibatch epochs          & 5 \\
value-loss coefficient    & 1.0 \\
entropy coefficient       & 0 \\
actor bounds-loss coeff.  & 0.01 \\
\bottomrule
\end{tabular}
\end{table}

\subsection{Damping Randomization and Inference}
\label{app:dyn}

During training, the damping scale factor on the target object joint
is sampled uniformly from $[1.0, 2.0]$ by default. Extended damping
experiments replace this interval with $[1.0, 4.0]$, $[1.0, 2.5]$,
or $[2.0, 4.0]$ to analyze the boundary of training-distribution
variation; friction randomization is not enabled in the current
experiments.

At evaluation, the model checkpoint is fixed and rollouts are
conducted at $\times1$, $\times2$, and $\times4$ damping. Deterministic execution
uses the policy mean and stochastic execution samples from the
learned Gaussian policy. Each quantitative cell uses 20 episodes by
default; single-rollout visualizations are reserved for behavioral
diagnostics and are not pooled into the multi-episode statistics.
Checkpoint selection should not rely on training reward alone but
should combine OOD damping evaluation, the action-saturation ratio,
and the detachment-failure rate.

\end{document}